# Learning Tensors in Reproducing Kernel Hilbert Spaces with Multilinear Spectral Penalties

Marco Signoretto[1], Lieven De Lathauwer[2], and Johan A.K. Suykens[1]

[1]ESAT-STADIUS, Katholieke Universiteit Leuven, Kasteelpark Arenberg 10, B-3001 Leuven (BELGIUM)

[2] Group Science, Engineering and Technology, Katholieke Universiteit Leuven, Campus Kortrijk, E. Sabbelaan 53, 8500 Kortrijk (BELGIUM)

**Abstract**

We present a general framework to learn functions in tensor product reproducing kernel Hilbert spaces (TP-RKHSs). The methodology is based on a novel representer theorem suitable for existing as well as new spectral penalties for tensors. When the functions in the TP-RKHS are defined on the Cartesian product of finite discrete sets, in particular, our main problem formulation admits as a special case existing tensor completion problems. Other special cases include transfer learning with multimodal side information and multilinear multitask learning. For the latter case, our kernel-based view is instrumental to derive nonlinear extensions of existing model classes. We give a novel algorithm and show in experiments the usefulness of the proposed extensions.

## 1 Introduction

Recently there has been an increasing interest in the cross-fertilization of ideas coming from kernel methods and tensor-based data analysis. On the one hand it became apparent that machine learning algorithms can greatly benefit from the rich structure of tensor-based data representations [53, 42, 43, 24]. In a distinct (but not unrelated) second line of research, parametric models consisting of higher order tensors made their way in inductive as well as transductive learning methods [44, 39, 51]. Within transductive techniques, in particular, *tensor completion* and *tensor recovery* emerged as a useful higher order generalization of their matrix counterpart [30, 19, 45, 44, 49, 33, 29, 10, 31]. By dealing with the estimation of tensors in the unifying framework of reproducing kernel Hilbert spaces (RKHSs), this paper positions itself within this second line of research. More precisely, we study penalized empirical risk minimization for learning tensor product functions. We propose a new class of regularizers, termed *multilinear spectral penalties*, that is related to spectral regularization for operator estimation [1]. Kernel methods for function estimation using a quadratic penalty and a tensor product kernel (such as the Gaussian-RBF kernel) arise as a special case. More interestingly, when the functions in the RKHS are defined on the Cartesian product of finite discrete sets, our formulation specializes to existing tensor completion problems. Other special cases include transfer learning with multimodal side information and multilinear multitask learning, recently proposed in [39]. We show that the problem formulations studied in [39] are equivalent to special instances of the formulation studied here. In turn, whereas the formulations in [39] give rise to models that are linear in the data, our kernel-based interpretation enables for natural nonlinear extensions, which are proven useful in experiments.

The concept of tensor product function is far from new. Within nonparametric statistics, in particular, tensor product splines have found application in density estimation and in the approximation of multivariate functions [50, 22, 18, 26]. In machine learning, tensor product has been used to build kernels for a long time [20]. The widely used Gaussian-RBF kernel, in particular, is an important example. However, the structure of tensor product functions has not been exploited in the existing regularization mechanisms. The class of multilinear spectral penalties proposed in this paper goes in the direction of filling this gap. The approach is naturally related to a notion of multilinear rank for abstract tensors that can be found in [23] in the context of Banach spaces. Here we focus on reproducing kernel Hilbert spaces which allows us to conveniently formulate problems of learning from data. Our approach relies on the generalization to the functional setting of a number of mathematical tools for finite dimensional tensors. In particular, central to our definition of multilinear spectral penalties is the notion of functional unfolding; the latter generalizes the idea of matricization for finite dimensional tensors. Remarkably, a number of penalties proposed in the literature are special cases of the class of regularizers that we study. We use this class of penalties to formulate a general penalized empirical risk minimization problem which comprises, as special cases, the learning frameworks mentioned before. We show that finding a solution to such a problem boils down to computing a finite dimensional tensor. We then focus on a penalty that combines the sum of nuclear norms of the functional unfoldings with an upper bound on the multilinear rank; we show that a restatement of a problem formulation employing the quadratic loss can be tackled by a simple block descent algorithm.

The paper is structured as follows. In the next section we introduce the concept of tensor product reproducing kernel Hilbert spaces together with instrumental tools for tensor product functions. In Section 3 we elaborate on multilinear spectral regularization, we formulate the main learning problem and discuss special cases. Section 4 is devoted to the characterization of solutions. In particular, we give a novel representer theorem suitable for general multilinear spectral penalties and elaborate on out of sample evaluations. In Section 5 we discuss algorithmical aspects and present experimental results in Section 6. We end the paper with concluding remarks in Section 7.

# 2 Tensor Product Reproducing Kernel Hilbert Spaces

## 2.1 Preliminaries

For a positive integer $I$ we denote by $[I]$ the set of integers up to and including $I$. We write $\times_{m=1}^{M}[I_m]$ to mean $[I_1] \times [I_2] \times \cdots \times [I_M]$, i.e., the cartesian product of such index sets, the elements of which are $M-$tuples $(i_1, i_2, \ldots, i_M)$. For a generic set $\mathcal{X}$, we write $\mathbb{R}^{\mathcal{X}}$ to mean the set of mappings from $\mathcal{X}$ to $\mathbb{R}$. Key to this work is to consider on the same footing functions that are defined on *discrete* and *continuous* domains. In particular, we will regard the set of real vectors (denoted by lower case letters $a, b, \ldots$) as a set of real-valued *functions*. Indeed a function $a : [I_1] \to \mathbb{R}$ corresponds to a $I_1-$dimensional vector. Likewise, matrices (denoted by bold-face capital letters $\boldsymbol{A}, \boldsymbol{B}, \ldots$) will be regarded as real-valued functions on the cartesian product of two finite index sets. Indeed note that the function $\boldsymbol{A} : [I_1] \times [I_2] \to \mathbb{R}$ corresponds to a $I_1 \times I_2$ matrix. We will write $\boldsymbol{A}_{\cdot n}$ (respectively, $\boldsymbol{A}_{n\cdot}$) to mean the $n$th column (respectively, row) of $\boldsymbol{A}$. Before proceeding, we recall here the notion of *Reproducing kernel Hilbert space*. The reader is referred to [6] for a detailed account. Let $(\mathcal{H}, \langle\cdot,\cdot\rangle)$ be a Hilbert space (HS) of real-valued functions on some set $\mathcal{X}$. A function $k : \mathcal{X}\times\mathcal{X} \to \mathbb{R}$ is said to be the *reproducing kernel* of $\mathcal{H}$ if and only if [4]:

a. $k(\cdot, x) \in \mathcal{H},\ \forall x \in \mathcal{X}$

b. $\langle f, k(\cdot, x)\rangle = f(x)\ \forall x \in \mathcal{X},\ \forall f \in \mathcal{H}$ (*reproducing property*).

In the following we often denote by $k_x$ the function $k(\cdot, x) : t \mapsto k(t, x)$. A HS of functions $(\mathcal{H}, \langle\cdot,\cdot\rangle)$ that possesses a reproducing kernel $k$ is a Reproducing kernel Hilbert space (RKHS); we denote it by $(\mathcal{H}, \langle\cdot,\cdot\rangle, k)$.

## 2.2 Tensor Product Spaces and Partial Tensor Product Spaces

Now assume that $\left(\mathcal{H}_m, \langle\cdot,\cdot\rangle_m, k^{(m)}\right)$ is a RKHS of functions defined on a domain $\mathcal{X}_m$, for any $m \in [M]$. We are interested in functions of the generic tuple

$$x = \left(x^{(1)}, x^{(2)}, \ldots, x^{(M)}\right) \in \mathcal{X},\ \text{ where } \mathcal{X} := \times_{m=1}^{M} \mathcal{X}_m\ .$$

We denote by $x^{(\beta)}$ the restriction of $x$ to $\beta \subseteq [M]$, i.e., $x^{(\beta)} = \left(x^{(m)}\ :\ m \in \beta\right)$. In the following we consider the vector space formed by the linear combinations of the functions $\otimes_{m\in\beta} f^{(m)}$, defined by:

$$\otimes_{m\in\beta} f^{(m)} : x^{(\beta)} \ \mapsto\ \prod_{m\in\beta} f^{(m)}\left(x^{(m)}\right)\ , \qquad f^{(m)} \in \mathcal{H}_m \text{ for any } m \in \beta\ . \tag{1}$$

The completion of this space according to the inner product:

$$\left\langle \otimes_{m\in\beta} f^{(m)}, \otimes_{m\in\beta} g^{(m)}\right\rangle_\beta := \prod_{m\in\beta} \left\langle f^{(m)}, g^{(m)}\right\rangle_m \tag{2}$$

gives the tensor product HS of interest, denoted by $\otimes_{m\in\beta}\mathcal{H}_m$. Whenever $\beta \subset [M]$ we call the space *partial* and denote it by[1] $(\boldsymbol{\mathcal{H}}_\beta, \langle\cdot,\cdot\rangle_\beta)$. When $\beta = [M]$ we write simply $(\boldsymbol{\mathcal{H}}, \langle\cdot,\cdot\rangle)$, i.e., we omit the $\beta$. In the rest of this paragraph we refer to this case, without loss of generality. In the following, $f^{(1)}\otimes f^{(2)}\otimes\cdots\otimes f^{(M)}$ or simply $\otimes_{m=1}^{M} f^{(m)}$, for compactness, is used to denote an elementary tensor[2] of the type (1), i.e., a *rank-1 tensor*. General elements of $\boldsymbol{\mathcal{H}}$ will be denoted by small bold type letters $(\boldsymbol{f}, \boldsymbol{g}, \ldots)$. Note that, by construction, $\boldsymbol{f} \in \boldsymbol{\mathcal{H}}$ can be expressed as the linear combination of rank-1 tensors:

$$\boldsymbol{f} = \sum_{\boldsymbol{\nu}} \alpha_{\boldsymbol{\nu}}\, f_{\nu_1}^{(1)} \otimes f_{\nu_2}^{(2)} \otimes \cdots \otimes f_{\nu_M}^{(M)} \tag{3}$$

where $\boldsymbol{\nu}$ is a multi-index, $\boldsymbol{\nu} = (\nu_1, \nu_2, \cdots, \nu_M)$. Consider now the symmetric function:

$$\boldsymbol{k} : (x, y) \mapsto k^{(1)}\left(x^{(1)}, y^{(1)}\right) k^{(2)}\left(x^{(2)}, y^{(2)}\right) \cdots\ k^{(M)}\left(x^{(M)}, y^{(M)}\right)\ . \tag{4}$$

It is easy to see that, for any $x \in \mathcal{X}$, $\boldsymbol{k}_x$ is a rank-1 tensor that belongs to $\boldsymbol{\mathcal{H}}$. Additionally one has:

$$\boldsymbol{f}(x) = \sum_{\boldsymbol{\nu}} \alpha_{\boldsymbol{\nu}} \left(f_{\nu_1}^{(1)} \otimes\cdots\otimes f_{\nu_m}^{(M)}\right)(x) = \sum_{\boldsymbol{\nu}} \alpha_{\boldsymbol{\nu}} \prod_{m=1}^{M} f_{\nu_m}^{(m)}\left(x^{(m)}\right) = \sum_{\boldsymbol{\nu}} \alpha_{\boldsymbol{\nu}} \prod_{m=1}^{M} \left\langle f_{\nu_m}^{(m)}, k^{(m)}\left(\cdot, x^{(m)}\right)\right\rangle_m = \langle \boldsymbol{f}, \boldsymbol{k}_x\rangle \tag{5}$$

[1] Throughout the paper we will use bold-face letters to denote tensor product spaces and functions.

[2] Note that whenever $\mathcal{X} = [I_1] \times [I_2] \times \cdots \times [I_M]$, in particular, $\otimes_{m=1}^{M} f^{(m)}$ corresponds to a finite dimensional higher order array.

which shows that $\boldsymbol{k}$ is the reproducing kernel of $\boldsymbol{\mathcal{H}}$. We call $\boldsymbol{\mathcal{H}}$ a *Tensor Product Reproducing Kernel Hilbert Spaces* (TP-RKHS) and denote it by $(\boldsymbol{\mathcal{H}}, \langle\cdot,\cdot\rangle, \boldsymbol{k})$. The norm induced by the inner product is $\|\boldsymbol{f}\| := \sqrt{\langle \boldsymbol{f}, \boldsymbol{f}\rangle}$. The generic partial space will be denoted by $(\boldsymbol{\mathcal{H}}_\beta, \langle\cdot,\cdot\rangle_\beta, \boldsymbol{k}^{(\beta)})$, with obvious meaning of the symbols. We have $\|\boldsymbol{f}\|_\beta := \sqrt{\langle \boldsymbol{f}, \boldsymbol{f}\rangle_\beta}$. Finally, for any $m \in [M]$ we will call $k^{(m)}$ a *factor kernel* and $\mathcal{H}_m$ a *factor space.*

## 2.3 Functional Unfolding and $m-$mode Product

For arbitrary $\beta \subset [M]$ we denote by $\beta^c$ its complement, i.e., $\beta^c := [M] \setminus \beta$. To each pair of vectors $(\boldsymbol{f}, \boldsymbol{g}) \in \boldsymbol{\mathcal{H}}_\beta \times \boldsymbol{\mathcal{H}}_{\beta^c}$ there corresponds a rank-1 operator $\boldsymbol{f} \otimes \boldsymbol{g} : \boldsymbol{\mathcal{H}}_{\beta^c} \to \boldsymbol{\mathcal{H}}_\beta$ defined by

$$(\boldsymbol{f} \otimes \boldsymbol{g})\boldsymbol{z} := \langle \boldsymbol{g}, \boldsymbol{z}\rangle_{\beta^c} \boldsymbol{f} \ . \tag{6}$$

One can show that the vector space generated by rank-1 operators equipped with the Hilbert-Frobenius inner product

$$\left\langle \boldsymbol{f}^{(1)} \otimes \boldsymbol{g}^{(1)}, \boldsymbol{f}^{(2)} \otimes \boldsymbol{g}^{(2)} \right\rangle_{\mathrm{HF}} := \left\langle \boldsymbol{f}^{(1)}, \boldsymbol{f}^{(2)} \right\rangle_\beta \left\langle \boldsymbol{g}^{(1)}, \boldsymbol{g}^{(2)} \right\rangle_{\beta^c} \tag{7}$$

forms the HS of so called *Hilbert-Schmidt operators* [14]. The *functional unfolding operator*[3] $M_\beta$, is now defined by:

$$M_\beta : \ \otimes_{m=1}^M f^{(m)} \ \mapsto \ \boldsymbol{f}^{(\beta)} \otimes \boldsymbol{f}^{(\beta^c)} \quad \text{with} \quad \begin{array}{rcl} \boldsymbol{f}^{(\beta)} & = & \otimes_{m\in\beta} f^{(m)} \in \boldsymbol{\mathcal{H}}_\beta, \\ \boldsymbol{f}^{(\beta^c)} & = & \otimes_{m\in\beta^c} f^{(m)} \in \boldsymbol{\mathcal{H}}_{\beta^c} \ . \end{array} \tag{8}$$

In particular, we write $M_j$ to mean the $j-$*mode unfolding* $M_{\{j\}}$. The functional unfolding operator can be regarded as a generalization of the matricization usually applied to finite dimensional tensors. In particular, the $j-$mode unfolding used, e.g., in [13, Definition 1] is equivalent to the definition given here. A closely related concept in the more general context of Banach spaces is found in [23, Definition 5.3]. As before, although (8) is stated for rank-1 tensors, the definition extends by linearity to a generic $\boldsymbol{f} \in \boldsymbol{\mathcal{H}}$. Note that $M_\beta$ is an isometry between the space $\boldsymbol{\mathcal{H}}$ and the space of Hilbert-Schmidt operators. In particular, it follows by the definition of inner products given above that, if $\boldsymbol{g}$ and $\boldsymbol{f}$ are rank-1 tensors, then:

$$\langle \boldsymbol{g}, \boldsymbol{f}\rangle = \langle M_\beta(\boldsymbol{g}), M_\beta(\boldsymbol{f})\rangle_{\mathrm{HF}} = \left\langle \boldsymbol{g}^{(\beta)} \otimes \boldsymbol{g}^{(\beta^c)}, \boldsymbol{f}^{(\beta)} \otimes \boldsymbol{f}^{(\beta^c)} \right\rangle_{\mathrm{HF}} \ . \tag{9}$$

Recall that, given arbitrary HSs $(\mathcal{F}, \langle\cdot,\cdot\rangle_{\mathcal{F}})$ and $(\mathcal{G}, \langle\cdot,\cdot\rangle_{\mathcal{G}})$, the adjoint of an operator $A : \mathcal{F} \to \mathcal{G}$ is the operator $A^* : \mathcal{G} \to \mathcal{F}$, satisfying $\langle A(f), g\rangle_{\mathcal{G}} = \langle f, A^*(g)\rangle_{\mathcal{F}}$. From this and the definitions of inner products (2) and (7) it follows that

$$M_\beta^* : \boldsymbol{f}^{(\beta)} \otimes \boldsymbol{f}^{(\beta^c)} \mapsto \otimes_{m=1}^M f^{(m)} \ ,$$

i.e., the adjoint $M_\beta^*$ acts on rank-1 operators by simply "reversing" the factorization.

We conclude this section with one additional definition which is usually stated for finite dimensional tensors (see e.g. [13, Definition 8]). For $m \in \beta \subseteq [M]$ consider the linear operator $A^{(m)} : \mathcal{H}_m \to \mathcal{G}_m$ where $\mathcal{G}_m$ denotes a HS. The $m-$mode product between $\boldsymbol{f}$ and $A^{(m)}$ is that element of $\mathcal{H}_1 \otimes \mathcal{H}_2 \otimes \cdots \otimes \mathcal{H}_{m-1} \otimes \mathcal{G}_m \otimes \mathcal{H}_{m+1} \otimes \cdots \otimes \mathcal{H}_M$ implicitly defined by:

$$\boldsymbol{g} = \boldsymbol{f} \times_m A^{(m)} \quad \Leftrightarrow \quad M_m(\boldsymbol{g}) = A^{(m)} M_m(\boldsymbol{f}) \ . \tag{10}$$

[3]We refer to $M_\beta$ as the functional unfolding operator; we call $M_\beta(\boldsymbol{f})$ the ($\beta-$)functional unfolding of $\boldsymbol{f}$.

An explicit definition as well as properties are given in Appendix B in connection to the Kronecker product of operators. The material presented so far will suffice for the study of penalized tensor estimation problems in the upcoming sections. In Appendix A we collected additional notes useful for the implementation of Algorithms.

# 3 Learning Tensors in RKHSs

## 3.1 Multilinear Spectral Regularization

### 3.1.1 Multilinear Rank

Our interest in functional unfoldings arises from the following definition of $\beta-$rank:

$$\text{rank}_\beta(\boldsymbol{f}) := \dim\left\{(M_\beta(\boldsymbol{f}))\,\boldsymbol{z}\ :\ \boldsymbol{z}\in\boldsymbol{\mathcal{H}}_{\beta^c}\right\} \tag{11}$$

which can be found in [23, Chapter 5.2] for general infinite dimensional spaces. In the following, on the other hand, we focus on learning from data. Our interest is on approaches that seek for predictive models spanning minimal subspaces along different functional unfoldings. Let $\mathcal{B}$ be a partition of $[M]$; our notion of model complexity relates to the following tuple:

$$\text{mlrank}_{\mathcal{B}}(\boldsymbol{f}) := \left(\text{rank}_\beta(\boldsymbol{f})\ :\ \beta\in\mathcal{B}\right)\ . \tag{12}$$

A particular instance of the former is the $M-$tuple:

$$\text{mlrank}(\boldsymbol{f}) := \left(\text{rank}_{\{1\}}(\boldsymbol{f}), \text{rank}_{\{2\}}(\boldsymbol{f}), \ldots, \text{rank}_{\{M\}}(\boldsymbol{f})\right) \tag{13}$$

which is usually called *multlinear rank*, in a finite dimensional setting [13]. By extension, we call (12) the $\mathcal{B}-$multilinear rank. In the following we denote by $\boldsymbol{\mathcal{H}}_\beta\otimes\boldsymbol{\mathcal{H}}_{\beta^c}$ the set of *compact operators*[4] from $\boldsymbol{\mathcal{H}}_{\beta^c}$ to $\boldsymbol{\mathcal{H}}_\beta$. If $M_\beta(\boldsymbol{f})\in\boldsymbol{\mathcal{H}}_\beta\otimes\boldsymbol{\mathcal{H}}_{\beta^c}$, then the following *spectral decomposition* holds (see, e.g., [23, Theorem 4.114]):

$$M_\beta(\boldsymbol{f}) = \sum_r \sigma_r^{(\beta)}\boldsymbol{u}_r\otimes\boldsymbol{v}_r\ , \tag{14}$$

with orthonormal families $\{\boldsymbol{u}_r\}_r\subset\boldsymbol{\mathcal{H}}_\beta$, $\{\boldsymbol{v}_r\}_r\subset\boldsymbol{\mathcal{H}}_{\beta^c}$ and scalars $\sigma_1^{(\beta)}\geq\sigma_2^{(\beta)}\geq\cdots\geq 0$ with $\lim_{i\to\infty}\sigma_i^{(\beta)}=0$. We can then relate the notion of rank in (11) to (14) by:

$$\text{rank}_\beta(\boldsymbol{f}) = \min\left\{r\ :\ \sigma_{r+1}(M_\beta(\boldsymbol{f}))=0\right\}\quad\text{where}\quad\sigma_r(M_\beta(\boldsymbol{f})) := \sigma_r^{(\beta)} \tag{15}$$

and we allow $\text{rank}_\beta(\boldsymbol{f})$ to take value infinity. Examples of finite dimensional tensors with given multilinear rank, which arise as a special case of the present framework, can be found in the literature on tensor-based methods. We will further discuss the finite dimensional case later, in connection to completion problems. The next example, on the other hand, fully relies on the functional tools introduced so far.

[4]The class of compact operators represents a natural generalization to the infinite-dimensional setting of the class of linear operators between finite dimensional spaces. Technically, compact operators on HSs are limits (in the operator norm) of sequences of finite-rank operators see, e.g., [11] for an introduction.

### 3.1.2 Tensor Product Functions and Multilinear Rank: an Example

As an illustration of the concepts introduced above, consider the function $\boldsymbol{f} : [0, 2\pi] \times [0, 2\pi] \times [0, 2\pi] \to \mathbb{R}$ defined by:

$$\boldsymbol{f}(x) := 2\sin\left(x^{(1)}\right) + \sin\left(2x^{(2)}\right) + 3\sin\left(x^{(2)}\right)\sin\left(4x^{(3)}\right) + \sin\left(x^{(1)}\right)\sin\left(x^{(3)}\right) \, . \tag{16}$$

We proceed by showing that the latter can be regarded as a tensor product function in an infinite dimensional RKHS; additionally we show that, within this space, we have mlrank$(\boldsymbol{f}) = (2, 3, 3)$. To see this, consider for $m \in \{1, 2, 3\}$ and $t \geq 0$, the HSs:

$$\mathcal{H}_m = \left\{ f^{(m)} \; : \; f^{(m)}\left(x^{(m)}\right) := \frac{a_0}{\sqrt{2}} + \sum_{l\in\mathbb{N}} a_l \psi_l^m\left(x^{(m)}\right), \; \left\|f^{(m)}\right\|_m := \frac{a_0^2}{2} + \sum_{l\in\mathbb{N}} l\exp(2lt)a_l^2 < \infty \right\}. \tag{17}$$

where $\psi_l : x \mapsto \sin(lx)$. It is not difficult to see that the function[5]:

$$k^{(m)}\left(x^{(m)}, y^{(m)}\right) := \frac{1}{2} + \sum_{l\in\mathbb{N}} \frac{\exp(-2lt)}{l} \psi_l^m\left(x^{(m)}\right)\psi_l^m\left(y^{(m)}\right) = \frac{1}{4}\ln\left(\exp(2)\frac{\sin^2\left(\frac{x^{(m)}+y^{(m)}}{2}\right) + \sinh^2(t)}{\sin^2\left(\frac{x^{(m)}-y^{(m)}}{2}\right) + \sinh^2(t)}\right) \tag{18}$$

is the reproducing kernel of $\mathcal{H}_m$ with respect to the inner product[6]

$$\langle f^{(m)}, g^{(m)} \rangle_m := \frac{1}{2}a_0 b_0 + \sum_{l\in\mathbb{N}} l\exp(2lt)a_l b_l$$

where $f^{(m)}$ is as in (17) and $g^{(m)} = \frac{b_0}{\sqrt{2}} + \sum_{l\in\mathbb{N}} b_l \psi_l^m\left(x^{(m)}\right)$. Now observe that, according to (17), we have

$$\left\{1, 2\psi_1^1\right\} \subset \mathcal{H}^{(1)}, \left\{1, 3\psi_1^2, \psi_2^2\right\} \subset \mathcal{H}^{(2)} \text{ and } \left\{1, 1/2\psi_1^3, \psi_4^3\right\} \subset \mathcal{H}^{(3)}$$

where, with some abuse of notation, we denoted by 1 the function identically equal to one. We can now equivalently restate $\boldsymbol{f}$ as the sum of rank-1 tensors[7]:

$$\boldsymbol{f} = 2\psi_1^1 \otimes 1 \otimes 1 + 1 \otimes \psi_2^2 \otimes 1 + 1 \otimes 3\psi_1^2 \otimes \psi_4^3 + 2\psi_1^1 \otimes 1 \otimes 1/2\psi_1^3 \tag{19}$$

which shows that $\boldsymbol{f}$ can be regarded as an element of $\boldsymbol{\mathcal{H}} := \mathcal{H}_1 \otimes \mathcal{H}_2 \otimes \mathcal{H}_3$. With reference to (6) we now have:

$$(M_1(\boldsymbol{f}))\boldsymbol{z} = \left\langle \psi_2^2 \otimes 1 + 3\psi_1^2 \otimes \psi_4^3, \boldsymbol{z} \right\rangle_{\{1\}^c} 1 + \left\langle 1 \otimes 1 + 1 \otimes 1/2\psi_1^3, \boldsymbol{z} \right\rangle_{\{1\}^c} 2\psi_1^1 \tag{20}$$

and therefore (11) gives rank$_1(\boldsymbol{f}) = 2$; in a similar fashion one obtains rank$_2(\boldsymbol{f}) = 3$ and rank$_3(\boldsymbol{f}) = 3$, i.e., mlrank$(\boldsymbol{f}) = (2, 3, 3)$. Next we show how to impose the structural information of low multilinear rank. In Section 6, we will show that this leads to significantly outperforming a learning algorithm that only relies on the smoothness of the generating function.

[5] Note that the last equality in (18) follows from a result on the series of trigonometric functions, see [21, Equation 1.462].

[6] That is, $k$ satisfies the properties $a$ and $b$ of Section 2.1.

[7] Note that the decomposition is not unique; nevertheless, all decompositions give rise to the same multilinear rank.

#### 3.1.3 Penalty Functions

Our next goal is to introduce a class of penalties that exploits in learning problems the structure of tensor product reproducing kernel Hilbert spaces. Denote by $\boldsymbol{f}$ a generic tensor belonging to such a space. In this paper we are interested in the case where the order of $\boldsymbol{f}$ (denoted as $M$) is at least three. For the sake of learning, in turn, this model will be regarded as a tensor of order $Q$, with $2 \leq Q \leq M$. There are many situations where one may be interested in the case where $Q \neq M$. For instance, consider the problem of approximating scattered data in $M$ dimensions by a tensor product function of order $M$. One might want to group variables into $Q$ disjoint groups and preserve this grouping in the regularization mechanism. A concrete example will be studied in Section 6.3 in connection to transfer learning with multimodal information (Section 3.3.3). We are now ready to introduce the class of penalty functions of interest, namely *multlilinear spectral penalties* (MSPs).

**Definition 1** ($\mathcal{B}$-MSP). *Let $\mathcal{B} = \{\beta_1, \beta_2, \ldots, \beta_Q\}$ be a partition of $[M]$ with cardinality $Q \geq 2$. For $s : \mathbb{R}^+ \times \mathbb{N} \times \mathbb{N}_Q \to \mathbb{R}^+$, we call*

$$\Omega_{\mathcal{B}}(\boldsymbol{f}) := \begin{cases} \sum_{q\in[Q]} \sum_{r\in\mathbb{N}} s\left(\sigma_r(M_{\beta_q}(\boldsymbol{f})), r, q\right), & \text{if } Q > 2 \\ \sum_{r\in\mathbb{N}} s\left(\sigma_r(M_{\beta_1}(\boldsymbol{f})), r, q\right), & \text{if } Q = 2 \end{cases} \tag{21}$$

*a $\mathcal{B}$-multilinear spectral penalty ($\mathcal{B}$-MSP) function if and only if:*

1. *$s(\cdot, r, q)$ is a nondecreasing function for any $(r, q) \in \mathbb{N} \times [Q]$*
2. *$s(0, r, q) = 0$ for any $(r, q) \in \mathbb{N} \times [Q]$.*

Notably, in light of (13) and (15), the function in (21) represents a natural extension towards tensors of the concept of spectral penalty function found for the two-way case in [1]. In the following, unless stated differently, we always assume that $Q > 2$. The case $Q = 2$, in turn, corresponds to the situation where the tensor is regarded as a matrix. Note that, in this case, one has:

$$\frac{1}{2}\sum_{r\in\mathbb{N}} \left(s\left(\sigma_r(M_{\beta_1}(\boldsymbol{f})), r, q\right) + s\left(\sigma_r(M_{\beta_2}(\boldsymbol{f})), r, q\right)\right) = \sum_{r\in\mathbb{N}} s\left(\sigma_r(M_{\beta_1}(\boldsymbol{f})), r, q\right) = \sum_{r\in\mathbb{N}} s\left(\sigma_r(M_{\beta_2}(\boldsymbol{f})), r, q\right) \tag{22}$$

which motivates the distinction between the two cases in (21).

In order to define the MSPs that we are mostly concerned with, we need to introduce the Schatten-$p$ norms for compact operators. For $p \geq 1$ we call the operator $A$ *$p$-summable* if it satisfies $\sum_{n\geq 1} \sigma_n(A)^p < \infty$ . Note that *finite rank operators*, i.e., operators that can be decomposed as the sum of finitely many rank-1 operators[8], are always $p-$summable (regardless of the value of $p$). For a $p$-summable operator we define the Shatten-$p$ norm:

$$\|A\|_p := \left(\sum_{n\geq 1} \sigma_n(A)^p\right)^{1/p} . \tag{23}$$

[8]Note that an operator can have finite rank even though it is defined between infinite dimensional spaces. Operators between finite dimensional spaces are always finite rank.

In particular, the Shatten-1 norm is called *nuclear* (or *trace*); 1−summable operators are called *trace-class*. The Shatten-2 norm corresponds to the *Hilbert-Frobenius* norm. It can be restated in term of the inner product (7) as $\|A\|_2 = \sqrt{\langle A, A\rangle_{\mathrm{HF}}}$; we have seen already that 2−summable operators are called Hilbert-Schmidt. We are now ready to introduce the MSPs that we are especially interested in:

- Taking $s(u, r, q) = u^2$ leads to the sum of Hilbert-Frobenius norms:

$$\Omega_{\mathcal{B}}(\boldsymbol{f}) = \begin{cases} \sum_{q\in[Q]} \|M_{\beta_q}(\boldsymbol{f})\|_2^2, & \text{if } M_{\beta_q}(\boldsymbol{f}) \text{ is } 2-\text{summable for any } \beta_q \in \mathcal{B} \\ \infty, & \text{otherwise .} \end{cases} \tag{24}$$

- Taking $s(u, r, q) = u$ leads to the sum of nuclear norms:

$$\Omega_{\mathcal{B}}(\boldsymbol{f}) = \begin{cases} \sum_{q\in[Q]} \|M_{\beta_q}(\boldsymbol{f})\|_1, & \text{if } M_{\beta_q}(\boldsymbol{f}) \text{ is } 1-\text{summable for any } \beta_q \in \mathcal{B} \\ \infty, & \text{otherwise .} \end{cases} \tag{25}$$

- For $p \in \{1, 2\}$ and $R_q \geq 1$ for any $q \in [Q]$, taking:

$$s(u, r, q) = \begin{cases} u^p, & \text{if } r \leq R_q \\ \infty, & \text{otherwise} \end{cases}$$

  leads to a mlrank-constrained variant of the MSPs in the previous bullets:

$$\Omega_{\mathcal{B}}(\boldsymbol{f}) = \begin{cases} \sum_{q\in[Q]} \|M_{\beta_q}(\boldsymbol{f})\|_p^p, & \text{if } \operatorname{rank}_{\beta_q}(\boldsymbol{f}) \leq R_q \ \forall\, q \in [Q] \\ \infty, & \text{otherwise .} \end{cases} \tag{26}$$

### 3.2 Main Problem Formulation

We consider supervised learning problems based on a dataset $\mathcal{D}_N$ of $N$ input-output training pairs:

$$\mathcal{D}_N := \left\{ \left( x_n^{(1)}, x_n^{(2)}, \ldots, x_n^{(M)}, y_n \right) \in (\mathcal{X}_1 \times \mathcal{X}_2 \times \cdots \times \mathcal{X}_M) \times \mathcal{Y} \ : \ n \in [N] \right\} \tag{27}$$

where $\mathcal{Y} \subseteq \mathbb{R}$ is the target space and $\mathcal{X} = \mathcal{X}_1 \times \mathcal{X}_2 \times \cdots \times \mathcal{X}_M$ is the input space. Different learning frameworks arise from different specifications of the latter. Some special cases are reported in Table 1; details are given in the next section.

Table 1: Different frameworks relate to different specifications of $\mathcal{X}$ in (27).

| *input space* $\mathcal{X}$ | *learning framework* |
|---|---|
| $[I_1] \times [I_2] \times \cdots \times [I_M]$ | finite dimensional tensor completion |
| $\mathcal{G} \times [I_2] \times \cdots \times [I_M]$ | multilinear multitask learning |
| $\mathcal{G}_1 \times \mathcal{G}_2 \times \cdots \times \mathcal{G}_M$ | transfer learning based on multimodal information |

For a generic TP-RKHS $(\boldsymbol{\mathcal{H}}, \langle\cdot,\cdot\rangle, \boldsymbol{k})$ of functions on $\mathcal{X}$, define now:

$$\boldsymbol{H}_{\mathcal{B}} := \{ \boldsymbol{f} \in \boldsymbol{\mathcal{H}} \ : \ M_\beta(\boldsymbol{f}) \in \boldsymbol{\mathcal{H}}_\beta \otimes \boldsymbol{\mathcal{H}}_{\beta^c} \text{ for any } \beta \in \mathcal{B} \} \ . \tag{28}$$

We are concerned with the following class of penalized empirical risk minimization problem:

$$\min_{\boldsymbol{f}\in\boldsymbol{H}_{\mathcal{B}}}\left\{\sum_{(x,y)\in\mathcal{D}_N} l\left(y,\langle\boldsymbol{f},\boldsymbol{k}_x\rangle\right)+\lambda\,\Omega_{\mathcal{B}}(\boldsymbol{f})\right\} \tag{29}$$

in which $l:\mathbb{R}\times\mathbb{R}\to\mathbb{R}$ is some loss function, $\lambda>0$ is a trade-off parameter and $\Omega_{\mathcal{B}}$ is a $\mathcal{B}$-MSP. In the following, unless stated differently, we do not assume that $l$ is convex. Note that the definition of $\boldsymbol{H}_{\mathcal{B}}$ ensures that the decomposition (14) exists for any $\beta\in\mathcal{B}$ and therefore that any specific instance of $\mathcal{B}$-MSP in Section 3.1.3 is well defined. Additionally, whenever the set $\{\boldsymbol{f}\in\boldsymbol{H}_{\mathcal{B}}\ :\ \Omega_{\mathcal{B}}(\boldsymbol{f})<\infty\}$ is not empty, any solution must satisfy $\Omega_{\mathcal{B}}(\boldsymbol{f})<\infty$. The general formulation in (29) can be specialized into a variety of different inference problems. Next we relate it to existing frameworks as well as novel extensions.

## 3.3 Some Special Cases

### 3.3.1 Convex Finite Dimensional Tensor Completion $[\mathcal{X}=[I_1]\times[I_2]\times\cdots\times[I_M]]$

When $\left(\mathcal{H}_m,\langle\cdot,\cdot\rangle_m,k^{(m)}\right)$ is a RKHS of functions on $[I_m]$, $\mathcal{H}_m$ corresponds to the space of $I_m-$dimensional vectors $\mathbb{R}^{[I_m]}$. In this setting, $\boldsymbol{\mathcal{H}}:=\mathbb{R}^{[I_1]}\otimes\mathbb{R}^{[I_2]}\otimes\cdots\otimes\mathbb{R}^{[I_M]}$ is identified with $\mathbb{R}^{[I_1]\times[I_2]\times\cdots\times[I_M]}$ and (1), in particular, corresponds to the outer-product of vectors $f^{(m)}\in\mathbb{R}^{[I_m]}$, for $m\in[M]$. If, for any $m\in[M]$, $\langle\cdot,\cdot\rangle_m$ is the canonical inner product in $\mathbb{R}^{I_m}$, then the reproducing kernel is $k^{(m)}(i_m,i'_m)=\delta(i_m-i'_m)$ and we have:

$$\boldsymbol{k}(\boldsymbol{i},\boldsymbol{i}')=\prod_{m\in[M]}\delta(i_m-i'_m) \tag{30}$$

where we denoted by $\delta$ the Dirac delta function, $\delta(x):=1$ if $x=0$ and $\delta(x):=0$ if $x\neq0$. Correspondingly, the evaluation functional $\boldsymbol{k}_{\boldsymbol{i}}$ is simply given by the outer-product of canonical basis vectors:

$$\boldsymbol{k}_{\boldsymbol{i}}=e^{(i_1)}\otimes e^{(i_2)}\otimes\cdots\otimes e^{(i_M)}\text{ where } e^{(i_m)}\in\mathbb{R}^{I_m}\ :\ \mathrm{e}_j^{(i_m)}=\begin{cases}1, & \text{if } j=i_m\\ 0, & \text{otherwise}\end{cases},\ m\in[M] \tag{31}$$

and the reproducing property for a rank-1 tensor reads:

$$\left\langle f^{(1)}\otimes f^{(2)}\otimes\cdots\otimes f^{(M)},\boldsymbol{k}_{\boldsymbol{i}}\right\rangle=\left(f^{(1)}\otimes f^{(2)}\otimes\cdots\otimes f^{(M)}\right)_{\boldsymbol{i}}=f^{(1)}_{i_1}f^{(2)}_{i_2}\cdots f^{(M)}_{i_M}\ . \tag{32}$$

Note that the generic input $x\in\mathcal{X}$ in (27) is here a multi-index $\boldsymbol{i}=(i_1,i_2,\ldots,i_M)$ representing the location at which the tensor $\boldsymbol{f}$ is observed.

**Sum of Nuclear Norm Approach** When $\mathcal{B}=\{\{1\},\{2\},\ldots,\{M\}\}$ and $s\,(u,r,q)=u$, the formulation in (29) leads to the estimation problems for finite dimensional tensors found in [30, 19, 44, 49] and later referred as the sum of nuclear norm (SNN) approach. In particular, the *indicator loss*:

$$l(a,b)=\begin{cases}0, & \text{if } a=b\\ \infty, & \text{otherwise}\end{cases} \tag{33}$$

leads to tensor hard-completion, as defined, e.g., in [44, Section 6.4]. In turn, this is a generalization of the constrained formulation solved by many matrix completion algorithms.

**Alternative Convex Relaxations** Recently it has been shown that the SNN approach can be substantially suboptimal [33, 40]. This result is related to the more general realization that using the sum of individual sparsity inducing norms is not always effective [34, 2]. To amend this problem, [33] proposed the *square norm* for finite dimensional tensors. The authors have shown that, for tensors of order $M \geq 4$, minimizing the square norm leads to improved recoverability conditions in comparison to the SNN approach. Notably, the square norm qualifies as a multilinear spectral penalty. Indeed, for $\mathcal{B} = \{[\lfloor M/2 \rfloor], [\lfloor M/2 \rfloor]^c\}$ and an element $\boldsymbol{f}$ of a TP-RKHS, the square norm can be restated as:

$$\Omega_{\mathcal{B}}(\boldsymbol{f}) := \begin{cases} \|M_{[\lfloor M/2 \rfloor]}(\boldsymbol{f})\|_1, & \text{if } M_{[\lfloor M/2 \rfloor]}(\boldsymbol{f}) \text{ is } 1-\text{summable} \\ \infty, & \text{otherwise} \end{cases} \tag{34}$$

where we denoted by $\lfloor s \rfloor$ the greatest integer lower-bound of $s$. This fact implies that the results of Section 4, and in particular Theorem 2, hold also for the square norm. Moreover, whereas the definition of square norm in [33] is given in the finite dimensional setting, (34) can be used to learn more general functions in TP-RKHSs. Yet another alternative penalty for finite dimensional tensors is given in [40]. However, this penalty does not belong to the class of multilinear penalties given in Definition 1.

### 3.3.2 Multilinear Multitask Learning $\left[\mathcal{X} = \mathcal{G} \times [I_2] \times \cdots \times [I_M]\right]$

Multi-task Learning (MTL) works by combining related learning tasks; it often improves over the case where tasks are learned in isolation, see for example [3, 5, 9] and references therein. Recently [39] has proposed an extension, termed Multilinear Multi-task Learning (MLMTL), to deal with the case where the structure of tasks is inherently multimodal. In the general case one may have input data consisting of elements of some metric space consisting, e.g., of (probability) distributions, graphs, dynamical systems, etc.; in the following we assume that such a space $\mathcal{G}$ has a Hilbert structure[9]. The approach in [39] only deals with the case where $\mathcal{G}$ is an Euclidean space $\mathbb{R}^{I_1}$ and only considers models that are linear in the data; their problem formulations work by estimating a finite dimensional tensor. Specifically, assume that there are $T$ linear regression tasks, each of which is represented by a vector $w_t \in \mathbb{R}^{I_1},\ t \in [T]$. In the case of interest $T = \prod_{m=2}^{M} I_m$ and the generic $t$ can be naturally represented as a multi-index $(i_2, i_3, \ldots, i_M)$; in this setting, the matrix obtained stacking the column vectors $w_t,\ t \in [T]$ is identified with the $1-$mode unfolding of a finite dimensional tensor $\boldsymbol{f} \in \mathbb{R}^{[I_1] \times [I_2] \times \cdots \times [I_M]}$:

$$[w_1, w_2, \ldots, w_T] = M_1(\boldsymbol{f}) \ . \tag{35}$$

Correspondingly, the data fitting term can be defined as[10]:

$$J(\boldsymbol{f}) := \sum_{t \in T} \sum_{(z,y) \in \mathcal{D}^{(t)}} l\big(y, w_t^\top z\big) \tag{36}$$

in which for any $t \in [T]$, $\mathcal{D}^{(t)} = \big\{ (z_n, y_n) \in \mathbb{R}^{I_1} \times \mathcal{Y} \ :\ n \in [N_t] \big\}$ is a task-dependent dataset. In order to encourage common structure among the tasks, [39] proposed two approaches; each of them can be seen as a special instance of the joint optimization problem:

$$\min_{\boldsymbol{f}} \{J(\boldsymbol{f}) + R(\boldsymbol{f})\} \tag{37}$$

[9] This is instrumental to encode in a simple manner (i.e., by means of a kernel function) a measure of similarity.
[10] This corresponds to equation 1 in [39].

in which $R(\boldsymbol{f})$ is a penalty that promotes low multilinear rank solutions. Interestingly, whenever $R$ is chosen to be a MSP, (37) can be shown to be equivalent to an instance of the general problem formulation in (29). This is an immediate consequence of the following proposition.

**Proposition 1.** *Denote by* $\kappa : [I_2] \times \cdots \times [I_M] \to [T]$ *a one-to-one mapping. For* $\mathcal{X} := \mathbb{R}^{[I_1]} \times [I_2] \times \cdots \times [I_M]$ *define*[11]*:*

$$\mathcal{D}_N := \left\{ \big(z, i_2, \ldots, i_M, y\big) \in \mathcal{X} \times \mathcal{Y} : \text{there exists } t = \kappa\big(i_2, \ldots, i_M\big) \text{ such that } (z, y) \in \mathcal{D}^{(t)} \right\} \ . \quad (38)$$

*Then if* $k : \mathcal{X} \times \mathcal{X} \to \mathbb{R}$ *is the tensor product kernel:*

$$\boldsymbol{k}\left(\big(z, i_2, \cdots, i_M\big), \big(z', i_2', \cdots, i_M'\big)\right) := z^\top z' \prod_{m=2}^{M} \delta\big(i_m - i_m'\big) \ , \quad (39)$$

*and* $\boldsymbol{f}$ *satisfies* (35)*, with reference to* (36) *it holds that:*

$$J(\boldsymbol{f}) = \sum_{(x,y) \in \mathcal{D}_N} l\left(y, \langle \boldsymbol{f}, \boldsymbol{k}_x \rangle\right) \ .$$

*Proof.* See Appendix C.1. □

This result reveals the underlying reproducing kernel (and the corresponding space of functions) associated to linear MLMTL in [39]. In light of this, it is now straightforward to generalize MLMTL to the case where models are not restricted to the linear case. Indeed one can replace $z^\top z'$ in (39) with a nonlinear kernel, such as the Gaussian-RBF:

$$\boldsymbol{k}\left(\big(z, i_2, \cdots, i_M\big), \big(z', i_2', \cdots, i_M'\big)\right) := \exp\left(-\|z - z'\|^2/\sigma^2\right) \prod_{m=2}^{M} \delta\big(i_m - i_m'\big) \ . \quad (40)$$

The next sections will show how to practically approach the problem in (29) to deal also with nonlinear MLMTL.

#### 3.3.3 Transfer Learning Based on Multimodal Information $\left[\mathcal{X} = \times_{m=1}^{M} \mathcal{G}_m\right]$

Generally speaking, transfer learning [35] and the related problem of collaborative filtering [25, 7] focus on applying knowledge gained while solving one problem to different but related problems. As a concrete example, consider the problem of learning user preferences ("Marco likes very much") on certain activity ("traveling") in different locations ("in the US"). In this case the preferences available at training can be regarded as values of entries in a third order finite dimensional tensor; users, activities and locations constitute the different modes of this tensor. If no side information were available, the task of finding missing entries (i.e., unobserved preferences in our running example) could be approached via tensor completion. We have seen, in particular, that standard tensor completion amounts at using a tensor product Dirac delta kernel (30). However, this approach makes it impossible to transfer knowledge to a new user and/or activity and/or location. In many practical problems, however, one additionally has multimodal side information at disposal on the entity types, each of which is identified with the mode of a tensor. For the entity user, for instance, one might have gender, age and occupation as attributes, see Figure 1 for an illustration.

[11] Note that $N$, the cardinality of $\mathcal{D}_N$, is by construction equal to $\prod_{t\in[T]} N_t$.

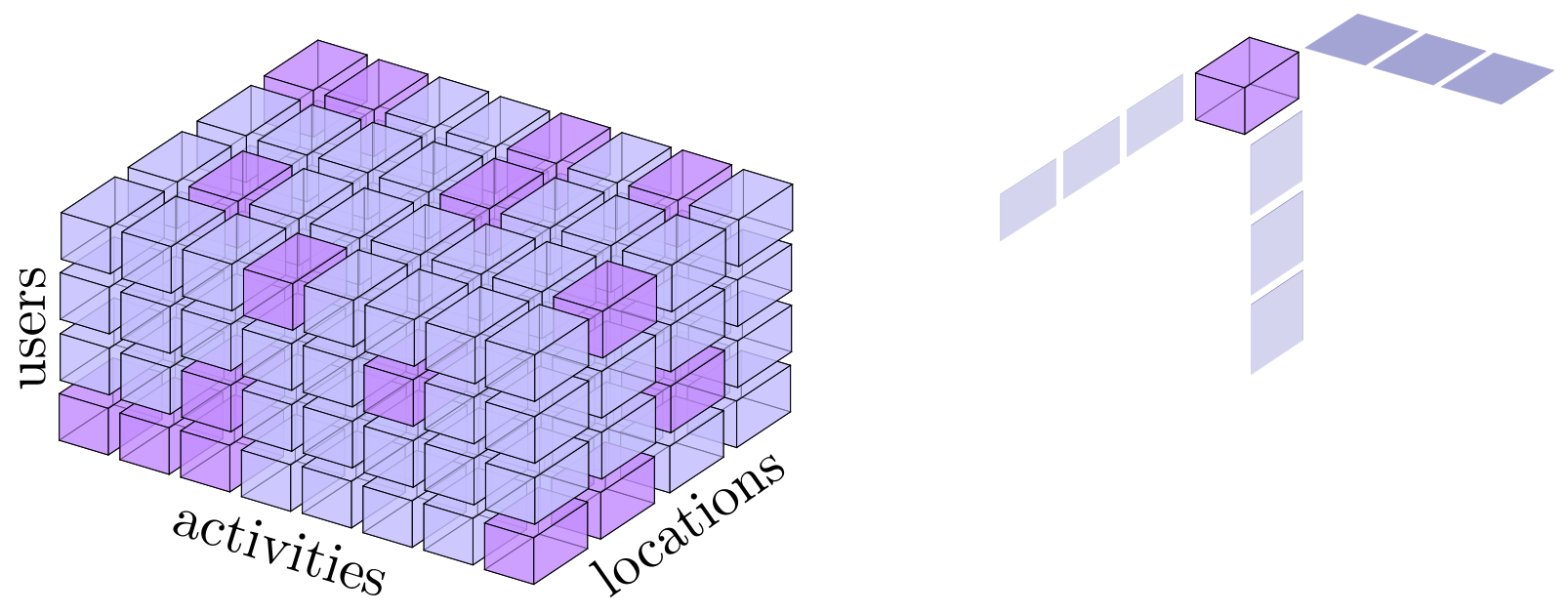

Figure 1: Learning preferences as a completion problem: the goal is to infer the value of unobserved entries (in dark purple) in a tensor with three modes, each of which corresponds to an entity type. Multimodal information available on each entry (e.g., a vector of attributes per mode) can be leveraged for the sake of transfer learning.

Mathematically speaking we can assume that, for the $m-$th mode, the information lies in some Hilbert space $\mathcal{G}_m$ (a finite dimensional feature space, a space of distributions, graphs, etc.). This space, in turn, is embedded into a Hilbert space of functions via a reproducing kernel $k^{(m)}$. This gives rise to a tensor product kernel of the type:

$$\boldsymbol{k}(\boldsymbol{i}, \boldsymbol{i}') = \prod_{m \in [M]} k^{(m)}\big(g^{(m)}_{i_m}, g^{(m)}_{i'_m}\big) \tag{41}$$

where $g^{(m)}_{i_m} \in \mathcal{G}_m$ (respectively $g^{(m)}_{i'_m} \in \mathcal{G}_m$) is associated to the $i_m$th (respectively, $i'_m$th) entity along the $m$-mode (a user, activity or location, depending on $m$). Training amounts at using this kernel within an instance of the problem in (29); the resulting estimated model $\hat{\boldsymbol{f}}$ can be used to computing preferences associated to users/activities/locations available at training. Tensor completion [30, 19, 44, 49] works under a low multilinear rank assumption on the finite dimensional tensor $\boldsymbol{f} : I_1 \times I_2 \times I_3 \to \mathbb{R}$. Correspondingly, transfer learning based on multimodal information works under a low multilinear rank assumption on the more general tensor product function $\boldsymbol{f} : \mathcal{G}_1 \times \mathcal{G}_2 \times \mathcal{G}_3 \to \mathbb{R}$. Given a new user indexed by $i_1^*$, in order to find his/her preferences one has simply to compute $\hat{\boldsymbol{f}}\big(g^{(1)}_{i_1^*}, g^{(2)}_{i_2}, g^{(3)}_{i_3}\big)$ where $(i_2, i_3)$ indexes the desired combination of activity and location.

# 4 Characterization of Solutions

## 4.1 Finite Dimensional Representation

For any $\boldsymbol{f} \in \boldsymbol{\mathcal{H}}$ and $\beta \in \mathcal{B}$, it follows from (9) that $\|M_\beta(\boldsymbol{f})\|_2^2 = \|\boldsymbol{f}\|^2$. In light of this, $\Omega_{\mathcal{B}}$ is defined upon $s\,(u, r, q) = u^2$, we have:

$$\Omega_{\mathcal{B}}(\boldsymbol{f}) = \sum_{\beta \in \mathcal{B}} \|M_\beta(\boldsymbol{f})\|_2^2 = |\mathcal{B}|\|\boldsymbol{f}\|^2 \tag{42}$$

where $|\mathcal{B}|$ is the cardinality of $\mathcal{B}$. Note that the specific composition of $\mathcal{B}$ does not matter in this case. Correspondingly, (29) boils down to a penalized learning problem with a tensor product kernel and a quadratic penalty; the latter is defined upon the RKHS norm. The case $l(a,b) = (a-b)^2$, in particular, gives rise to *regularization networks* [17]. This setting gives rise to the classical representer theorem.

**Theorem 1** (classical representer theorem)**.** *Any $\hat{\boldsymbol{f}}$ minimizing the regularized risk functional:*

$$\min_{\boldsymbol{f}\in \boldsymbol{H}_{\mathcal{B}}} \left\{ \sum_{n\in[N]} l\left(y_n, \langle \boldsymbol{k}_{x_n}, \boldsymbol{f}\rangle\right) + \lambda \sum_{\beta\in\mathcal{B}} \|M_\beta(\boldsymbol{f})\|_2^2 \right\} \tag{43}$$

*admits a representation of the form*

$$\hat{\boldsymbol{f}} = \sum_{n\in[N]} \alpha_n \boldsymbol{k}_{x_n} \tag{44}$$

*with $\alpha \in \mathbb{R}^N$.*

*Proof.* See Appendix C.2. ☐

We are now ready for the main result of this section.

**Theorem 2** (representer theorem for general MSPs)**.** *Let $\mathcal{B} = \{\beta_1, \beta_2, \ldots, \beta_Q\}$ be a partition of $[M]$ and let $\boldsymbol{H}_{\mathcal{B}} \subset \boldsymbol{\mathcal{H}} = \otimes_{m=1}^M \mathcal{H}_m$ be defined as in* (28)*. For any $q \in [Q]$, let*

$$\left\{ u_{i_q}^{(q)} \ : \ i_q \in [I_q] \right\} \tag{45}$$

*be an orthonormal basis for:*

$$\mathcal{V}_q := \operatorname{span}\left\{ k^{(\beta_q)}\left(x^{(\beta_q)}, \cdot\right) \ : \ (x,y) \in \mathcal{D}_N \right\} \ . \tag{46}$$

*Define $\Gamma(A,q) := \sum_{r\in\mathbb{N}} s\left(\sigma_r(A), r, q\right)$ where $s$ satisfies the assumptions in Definition 1. Consider the optimization problem:*

$$\min_{\boldsymbol{f}\in \boldsymbol{H}_{\mathcal{B}}} \mathcal{P}_\lambda^{\mathcal{H}}(\boldsymbol{f}) \quad \text{where} \quad \mathcal{P}_\lambda^{\mathcal{H}}(\boldsymbol{f}) := \sum_{n\in[N]} l\left(y_n, \langle \boldsymbol{k}_{x_n}, \boldsymbol{f}\rangle\right) + \lambda \sum_{q\in[Q]} \Gamma\left(M_{\beta_q}(\boldsymbol{f}), q\right) \ . \tag{47}$$

*If the set of solutions is non-empty then there exists $\hat{\boldsymbol{f}} \in \arg\min_{\boldsymbol{f}\in\boldsymbol{H}_{\mathcal{B}}} \mathcal{P}_\lambda(\boldsymbol{f})$, and $\hat{\boldsymbol{\alpha}} \in \mathbb{R}^{[I_1]\times[I_2]\times\cdots\times[I_Q]}$, such that the finite-rank tensor $\hat{\boldsymbol{g}} \in \boldsymbol{\mathcal{V}}_N := \otimes_{q=1}^Q \mathcal{V}_q$:*

$$\hat{\boldsymbol{g}} = \sum_{i_1\in[I_1]} \sum_{i_2\in[I_2]} \cdots \sum_{i_Q\in[I_Q]} \hat{\boldsymbol{\alpha}}_{i_1 i_2 \cdots i_Q} u_{i_1}^{(1)} \otimes u_{i_2}^{(2)} \otimes \cdots \otimes u_{i_Q}^{(Q)} \tag{48}$$

*satisfies:*

$$\hat{\boldsymbol{g}}\big(x^{(\beta_1)}, x^{(\beta_2)}, \ldots, x^{(\beta_Q)}\big) = \hat{\boldsymbol{f}}\big(x^{(1)}, x^{(2)}, \ldots, x^{(M)}\big) \ \ \forall \ \ \big(x^{(1)}, x^{(2)}, \ldots, x^{(M)}\big) \in \mathcal{X} \ . \tag{49}$$

*Proof.* See Appendix C.5. ☐

The result is related to [1, Theorem 3]. The latter deals with (compact) operators, which can be regarded as two-way tensors. In our result note that $Q$, the cardinality of the partition $\mathcal{B}$, is also the order of the tensor $\hat{\boldsymbol{g}}$ in (48). Since $Q \leq M$, where $M$ is the order of the tensors in $\boldsymbol{\mathcal{H}}$, we see that $\hat{\boldsymbol{g}}$ and $\hat{\boldsymbol{f}}$ might be tensors of different orders. In the general case, $\hat{\boldsymbol{g}}$ does not belong to $\boldsymbol{\mathcal{H}}$ and so $\hat{\boldsymbol{g}}$ is not a solution to (47), since $\boldsymbol{H}_{\mathcal{B}} \subset \boldsymbol{\mathcal{H}}$. Nonetheless, equation (49) states that we can use $\hat{\boldsymbol{g}}$ instead of $\hat{\boldsymbol{f}}$ for any practical purpose since the evaluations of $\hat{\boldsymbol{f}}$ and that of $\hat{\boldsymbol{g}}$ are identical. Our proof makes use of the following two instrumental results.

**Proposition 2.** *Let $\mathcal{B}$ and $\Gamma$ be as in Theorem 2; for any $q \in [Q]$ denote by $\mathcal{G}_q$ the tensor product space $\otimes_{m\in\beta_q}\mathcal{H}_m$. For an arbitrary $\gamma \subseteq [Q]$ define*[12] *$\boldsymbol{\mathcal{G}}_\gamma := \otimes_{q\in\gamma}\mathcal{G}_q$ and let:*

$$\boldsymbol{G}_{\mathcal{B}} := \left\{\boldsymbol{g} \in \boldsymbol{\mathcal{G}} \ : \ M_q(\boldsymbol{g}) \in \boldsymbol{\mathcal{G}}_{\{q\}} \otimes \boldsymbol{\mathcal{G}}_{\{q\}^c} \ \forall q \in [Q]\right\} \ . \tag{50}$$

*There is a vector space isomorphism $\iota : \boldsymbol{\mathcal{H}} \to \boldsymbol{\mathcal{G}}$ such that if $\hat{\boldsymbol{f}}$ is a solution to (47) then $\iota(\hat{\boldsymbol{f}})$ is a solution to:*

$$\min_{\boldsymbol{g}\in\boldsymbol{G}_{\mathcal{B}}} \mathcal{P}_\lambda^{\mathcal{G}}(\boldsymbol{g}) \quad \text{where} \quad \mathcal{P}_\lambda^{\mathcal{G}}(\boldsymbol{g}) := \sum_{n\in[N]} l\left(y_n, \langle \iota(\boldsymbol{k}_{x_n}), \boldsymbol{g}\rangle\right) + \lambda \sum_{q\in[Q]} \Gamma\left(M_q(\boldsymbol{g}), q\right) \ . \tag{51}$$

*vice-versa, if $\hat{\boldsymbol{g}}$ is a solution to (51) then $\iota^{-1}(\hat{\boldsymbol{g}})$ is a solution to (47). Moreover it holds that:*

$$\hat{\boldsymbol{g}}\big(x^{(\beta_1)}, x^{(\beta_2)}, \ldots, x^{(\beta_Q)}\big) = \iota^{-1}(\hat{\boldsymbol{g}})\big(x^{(1)}, x^{(2)}, \ldots, x^{(M)}\big) \quad \forall \ x = (x^{(1)}, x^{(2)}, \ldots, x^{(M)}) \in \mathcal{X} \ . \tag{52}$$

*Proof.* See Appendix C.3. □

**Lemma 1.** *Consider a generic tensor $\boldsymbol{g} \in \boldsymbol{\mathcal{G}} := \otimes_{q\in[Q]}\mathcal{G}_q$. For any $q \in [Q]$, denote by $\Pi_q$ the orthogonal projection onto a closed linear subspace of $\mathcal{G}_q$. For any $p \in [Q]$ it holds that:*

$$\sigma_r\big(M_p\big(\boldsymbol{g} \times_1 \Pi_1 \times_2 \Pi_2 \times_3 \cdots \times_Q \Pi_Q\big)\big) \leq \sigma_r\big(M_p(\boldsymbol{g})\big) \ \forall r \geq 1 \ . \tag{53}$$

*Proof.* See Appendix C.4. □

The following result shows that, even though $\hat{\boldsymbol{g}}$ might belong to an infinite dimensional space, it can be recovered based on finite dimensional optimization.

**Proposition 3.** *With reference to Theorem 2, one has:*

$$\hat{\boldsymbol{\alpha}} \in \arg \min_{\boldsymbol{\alpha}\in\mathbb{R}^{[I_1]\times[I_2]\times\cdots\times[I_Q]}} \left\{ \sum_{n\in[N]} l\left(y_n, \boldsymbol{\alpha} \times_1 \boldsymbol{F}_{n\cdot}^{(1)} \times_2 \cdots \times_Q \boldsymbol{F}_{n\cdot}^{(Q)}\right) + \lambda \sum_{q\in[Q]} \Gamma\left(M_q(\boldsymbol{\alpha}), q\right) \right\} \tag{54}$$

*where for any $q \in [Q]$, $\boldsymbol{F}^{(q)} \in \mathbb{R}^{[N]\times[I_q]}$ is a matrix that satisfies*

$$\boldsymbol{K}^{(q)} = \boldsymbol{F}^{(q)}\boldsymbol{F}^{(q)\top}, \ \text{where } \boldsymbol{K}_{ij}^{(q)} := k^{(\beta_q)}\left(x_i^{(\beta_q)}, x_j^{(\beta_q)}\right) \ \text{for } (x_i, y_i), (x_j, y_j) \in \mathcal{D}_N \ . \tag{55}$$

*Proof.* See Appendix C.6. □

[12] As before we write $\boldsymbol{\mathcal{G}}$ to mean $\otimes_{q\in[Q]}\mathcal{G}_q$.

A representation alternative to (48) is given in the proof of Proposition 3:

$$\hat{\boldsymbol{g}}(x) = \sum_{n_1\in[N]} \sum_{n_2\in[N]} \cdots \sum_{n_Q\in[N]} \boldsymbol{\gamma}_{n_1 n_2 \cdots n_Q} \boldsymbol{k}^{(\beta_1)} \left(x_{n_1}^{(\beta_1)}, x^{(\beta_1)}\right) \otimes \boldsymbol{k}^{(\beta_2)} \left(x_{n_2}^{(\beta_2)}, x^{(\beta_2)}\right) \otimes \cdots \otimes \boldsymbol{k}^{(\beta_Q)} \left(x_{n_Q}^{(\beta_Q)}, x^{(\beta_Q)}\right) \quad (56)$$

where $\boldsymbol{\gamma} \in \mathbb{R}^{[N]\times[N]\times\cdots\times[N]}$ is a $Q$th order tensor. Note that the representation in (44), valid for the case where the MSP is defined upon Schatten-2 norms, corresponds to the case where $\boldsymbol{\gamma}$ is the super-diagonal tensor:

$$\boldsymbol{\gamma}_{\boldsymbol{n}} = \begin{cases} \boldsymbol{\alpha}_n, & \text{if } \boldsymbol{n} = (n, n, \ldots, n) \\ 0, & \text{if } \boldsymbol{n} \neq (n, n, \ldots, n) \end{cases} .$$

The following result can be seen as the higher-order equivalent to [1, Corollary 4].

**Corollary 1.** *With reference to Theorem 2, suppose that, in problem* (47)*,* $\Omega_{\mathcal{B}}$ *is a mlrank-constrained MSP of the type* (68)*. Then the tensor* $\hat{\boldsymbol{\alpha}} \in \mathbb{R}^{[I_1]\times[I_2]\times\cdots\times[I_Q]}$ *satisfies:*

$$\mathrm{rank}_{\beta_q}(\hat{\boldsymbol{\alpha}}) \leq R_q, \ \textit{for any } q \in [Q] .$$

## 4.2 Out-of-sample Evaluations

Proposition 3 offers a recipe to find the parameter $\hat{\boldsymbol{\alpha}}$ within $\hat{\boldsymbol{g}}$ in (48). Still, in order to evaluate $\hat{\boldsymbol{g}}$ on a generic test point $x \in \mathcal{X}$ we need to be able to evaluate the orthogonal functions (45) on any point of the corresponding domains. In this section we illustrate how this can be done. For any $q \in [Q]$, denote by $u^{(q)} : \mathcal{X} \to \mathbb{R}^{I_q}$ the vector-valued function defined by $(u^{(q)}(x))_{i_q} := u_{i_q}^{(q)}\left(x^{(\beta_Q)}\right)$. Notice that, with reference to (48), the evaluation of $\hat{\boldsymbol{g}}$ on $x \in \mathcal{X}$ can be stated as:

$$\hat{\boldsymbol{g}}(x) = \hat{\boldsymbol{\alpha}} \times_1 u^{(1)\top}(x) \times_2 u^{(2)\top}(x) \times_3 \cdots \times_Q u^{(Q)\top}(x) \quad (57)$$

where $u^{(m)\top}(x)$ is the row vector obtained by evaluating $u^{(m)}$ at $x$. Now, by definition of $u_{i_q}^{(q)}$ we must have that:

$$(u^{(q)}(x))_{i_q} = \sum_{n\in[N]} (\boldsymbol{E}^{(q)})_{n i_q} k^{(\beta_q)}(x_n^{(\beta_q)}, x) \quad (58)$$

where we denoted by $\boldsymbol{E}^{(q)}$ the $N \times I_q$ matrix of coefficients for the vector-valued function $u^{(q)}$. In light of this, with reference to the generic training point $x_n,\ n \in [N]$, we obtain:

$$\hat{\boldsymbol{g}}(x_n) = \hat{\boldsymbol{\alpha}} \times_1 \boldsymbol{K}_{n\cdot}^{(1)} \boldsymbol{E}^{(1)} \times_2 \boldsymbol{K}_{n\cdot}^{(2)} \boldsymbol{E}^{(2)} \times_3 \cdots \times_Q \boldsymbol{K}_{n\cdot}^{(Q)} \boldsymbol{E}^{(Q)} . \quad (59)$$

On the other hand the evaluation of $\hat{\boldsymbol{g}}$ on $x_n$ can be stated also as:

$$\hat{\boldsymbol{g}}(x_n) = \hat{\boldsymbol{\alpha}} \times_1 \boldsymbol{F}_{n\cdot}^{(1)} \times_2 \cdots \times_Q \boldsymbol{F}_{n\cdot}^{(Q)} . \quad (60)$$

Therefore we see that $\boldsymbol{E}^{(q)}$ must satisfy:

$$\boldsymbol{K}^{(q)} \boldsymbol{E}^{(q)} = \boldsymbol{F}^{(q)} \quad (61)$$

$$\boldsymbol{E}^{(q)\top} \boldsymbol{K}^{(q)} \boldsymbol{E}^{(q)} = \boldsymbol{I} \quad (62)$$

where the second equation enforces that (45) is an orthonormal set, i.e., $\langle u_i^{(q)}, u_j^{(q)} \rangle_{\beta^q} = \delta(i-j)$. Keeping into account (55) it is not difficult to see that, if $\boldsymbol{A}^\ddagger$ denotes the transpose of the pseudo-inverse of a matrix $\boldsymbol{A}$, $\boldsymbol{E}^{(q)} = \boldsymbol{F}^{(q)\ddagger}$ is the unique solution to the system of equations (61)-(62). To conclude, we can evaluate $\hat{\boldsymbol{g}}$ on an arbitrary $x \in \mathcal{X}$ by:

$$\hat{\boldsymbol{g}}(x) = \hat{\boldsymbol{\alpha}} \times_1 \bar{k}^{(1)}(x)\boldsymbol{F}^{(1)\ddagger} \times_2 \bar{k}^{(2)}(x)\boldsymbol{F}^{(2)\ddagger} \times_3 \cdots \times_Q \bar{k}^{(Q)}(x)\boldsymbol{F}^{(Q)\ddagger} \quad \forall\, x \in \mathcal{X} \tag{63}$$

where for any $q \in [Q]$ we let:

$$\bar{k}^{(q)}(x) := \left[k^{(\beta_q)}\big(x_1^{(\beta_q)}, x^{(\beta_q)}\big),\ k^{(\beta_q)}\big(x_2^{(\beta_q)}, x^{(\beta_q)}\big), \ldots, k^{(\beta_q)}\big(x_N^{(\beta_q)}, x^{(\beta_q)}\big)\right] \ . \tag{64}$$

## 4.3 Link with Inductive Learning with Tensors

Recently, [44] has studied both transductive and inductive learning problems based on input data represented as finite dimensional tensors. For inductive learning the goal was to determine a model to be used for out of sample predictions. For single-task problems and no bias term, such a model can be stated as $\hat{m}(\boldsymbol{z}) = \langle \hat{\boldsymbol{\alpha}}, \boldsymbol{z} \rangle$ where $\boldsymbol{z}$ is a finite dimensional input data-tensor, and $\hat{\boldsymbol{\alpha}}$ is an estimated parameter, a tensor of the same dimensions as $\boldsymbol{z}$. With reference to (54), note that we have:

$$\boldsymbol{\alpha} \times_1 \boldsymbol{F}_{n\cdot}^{(1)} \times_2 \cdots \times_Q \boldsymbol{F}_{n\cdot}^{(Q)} = \langle \boldsymbol{\alpha}, \boldsymbol{z}_n \rangle \qquad \text{where} \qquad \boldsymbol{z}_n := \boldsymbol{F}_{n\cdot}^{(1)} \otimes \boldsymbol{F}_{n\cdot}^{(2)} \otimes \cdots \otimes \boldsymbol{F}_{n\cdot}^{(Q)} \ . \tag{65}$$

This shows that the finite dimensional problem (54) can be interpreted as an inductive learning problem with dataset:

$$\tilde{\mathcal{D}}_N := \left\{(\boldsymbol{z}_n, y_n) \in \mathbb{R}^{[I_1]\times[I_2]\times\cdots\times[I_Q]} \times \mathcal{Y} \ : \ n \in [N]\right\} \ , \tag{66}$$

see (27) for a comparison. Each input data $\boldsymbol{z}_n$ is a rank-1 tensor given by the outer-product of the vectors $\boldsymbol{F}_{n\cdot}^{(1)},\ \boldsymbol{F}_{n\cdot}^{(2)},\ \ldots,\ \boldsymbol{F}_{n\cdot}^{(Q)}$. In turn, each of these vectors is induced by one of the factor kernels giving rise to the tensor product kernel $\boldsymbol{k}$ in (4), see Figure 2 for an illustration.

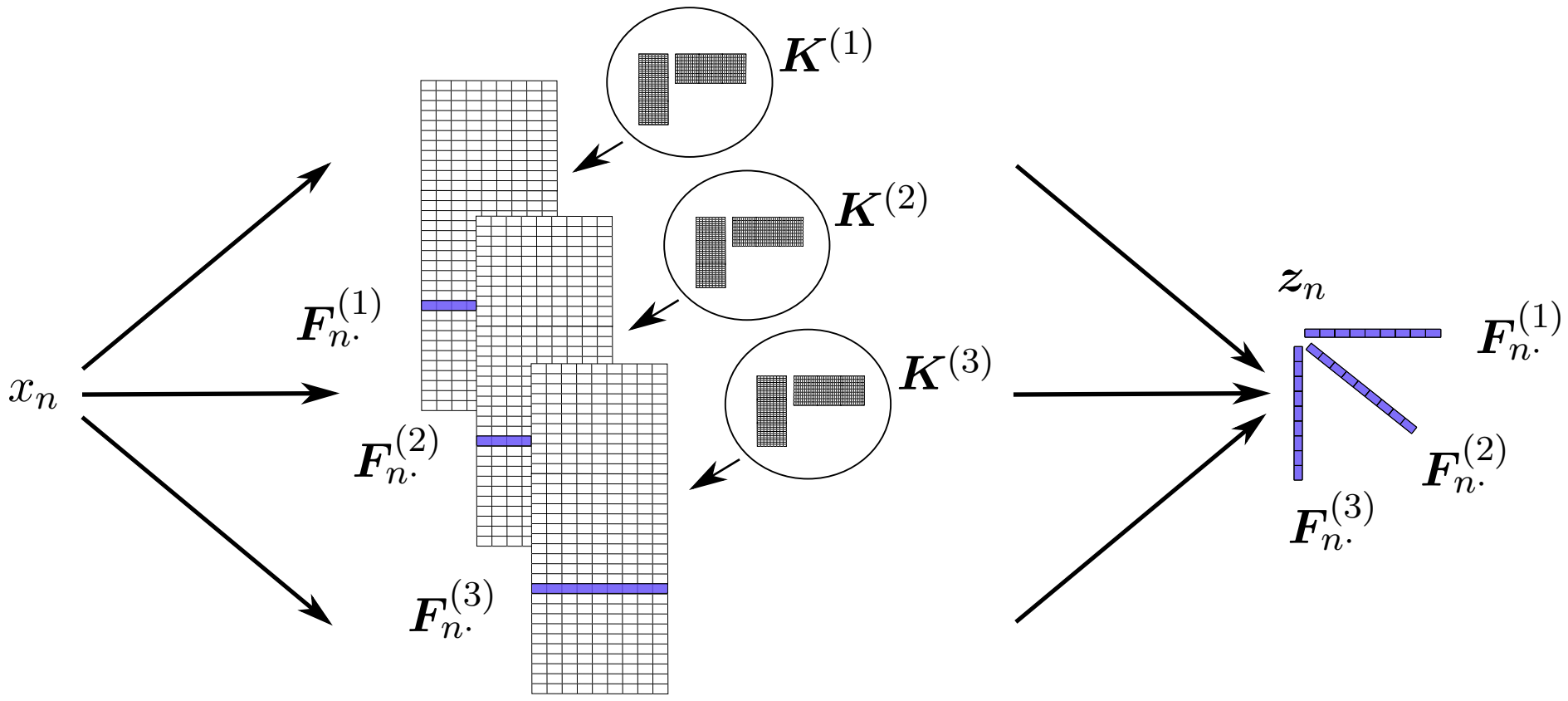

Figure 2: An illustration for $Q = 3$: each input data $\boldsymbol{z}_n$ is a rank-1 tensor.

The evaluation on a test point within the original domain $\mathcal{X}$ follows from (63):

$$x \mapsto \langle \hat{\boldsymbol{\alpha}}, \boldsymbol{z} \rangle \ : \ \boldsymbol{z} = \left(\bar{k}^{(1)}(x)\boldsymbol{F}^{(1)\ddagger}\right) \otimes \left(\bar{k}^{(2)}(x)\boldsymbol{F}^{(2)\ddagger}\right) \otimes \cdots \otimes \left(\bar{k}^{(Q)}(x)\boldsymbol{F}^{(Q)\ddagger}\right) \ . \tag{67}$$

## 5 Algorithmical Aspects

We have seen that solving the abstract problem formulation in (29) in practice amounts at finding a solution to (54). In turn, special instances of this finite dimensional optimization problem have been already studied in the technical literature and algorithms have been proposed. Our representer theorem implies that these algorithms can be easily adapted to find a tensor product function in a RKHS. In particular, in light of Section 4.3, one can directly rely on procedures specifically designed for inductive learning with tensors. More generally, we refer to [19, 44, 30, 52, 39, 33] for algorithms that deal with the case where the MSP is the sum of nuclear norm of the different matrix unfoldings. In the remainder of this section we focus on a different penalty that has not been explored so far for higher order tensors.

### 5.1 A Constrained Multilinear Rank Penalty

Here we consider the case where the MSP is a penalty obtained combining the sum of nuclear norms with an upper bound on the multilinear rank; the empirical risk, in turn, is measured by the quadratic loss. With reference to (29), we therefore have:

$$l(y,\hat{y}) = \frac{1}{2}\big(y-\hat{y}\big)^2 \qquad \text{and} \qquad \Omega_{\mathcal{B}}(\boldsymbol{f}) = \begin{cases} \sum_{q\in[Q]} \|M_{\beta_q}(\boldsymbol{f})\|_1, & \text{if } \operatorname{rank}_{\beta_q}(\boldsymbol{f}) \leq R_q \ \forall\, q \in [Q] \\ \infty, & \text{otherwise} \end{cases} \tag{68}$$

where $(R_1, R_2, \ldots, R_Q)$ is a user-defined tuple. The penalty that we consider is the natural higher order generalization of a regularization approach used for matrices in a number of different settings. In collaborative filtering, in particular, [38] has shown that the approach is suitable to tackle large-scale problems by gradient-based optimization; [15] has recently used a penalty based on the nuclear norm and a rank indicator function. The approach is used to find relevant low-dimensional subspaces of the output space. To accomplish this task one learns a low-rank kernel matrix; this is connected to a variety of methods, such as reduced-rank regression [27, 37]. In general, rank-constrained optimization problems are difficult non-convex problems with many local minima and only local search heuristics are known [46]. Nevertheless it has been recognized that setting an upper bound on the rank can be beneficial for both computational and statistical purposes. For certain matrix problems, if the upper bound is sufficiently large, one is guaranteed to recover the true solution under assumptions [8, 36]. For tensors, the empirical evidence in [39] suggests that a non-convex approach which specifies an explicit multilinear rank outperforms the convex SNN approach. In this respect note, however, that specifying the tuple $(R_1, R_2, \ldots, R_Q)$ in (68) is not an easy task. Indeed if we let $R_q \leq S$ for any $q \in [Q]$ and some global upper bound $S$, there are $S^Q$ different tuples to choose from. As in [39] our approach considers reduced tensors, which ensures reduced memory requirements and computational overhead. Our regularization mechanism, however, uses the nuclear norms to seek for a further reduction in the model $\boldsymbol{f}$. This allows us to pick a reasonable global upper bound $S$; we then take as upper-bound the tuple $(S, S, \ldots, S)$ and let the sum of nuclear norms enforce low dimensional subspaces along the different unfoldings. This results into a different regularization mechanism than the one in [39]. The finite dimensional optimization problem arising from (68) is:

$$\begin{aligned} \min_{\boldsymbol{\alpha}\in\mathbb{R}^{[I_1]\times[I_2]\times\cdots\times[I_Q]}} \quad & \tfrac{1}{2\lambda}\textstyle\sum_{n\in[N]} \Big(y_n - \boldsymbol{\alpha} \times_1 \boldsymbol{F}_{n\cdot}^{(1)} \times_2 \cdots \times_Q \boldsymbol{F}_{n\cdot}^{(Q)}\Big)^2 + \textstyle\sum_{q\in[Q]} \|M_q(\boldsymbol{\alpha})\|_1 \\ \text{subject to} \quad & \operatorname{rank}\big(M_q(\boldsymbol{\alpha})\big) \leq R_q \ \forall\, q \in [Q] \ . \end{aligned} \tag{69}$$

This problem formulation features a penalty on the SNN of the unfoldings of $\boldsymbol{\alpha}$ as well as an additional upper-bound on its multilinear rank; the latter results from the MSP imposed on the tensor product function $\boldsymbol{f}$, see Corollary 1. To approach problem (69) we consider the explicit Tucker parametrization [28]:

$$\boldsymbol{\alpha} = \boldsymbol{\beta} \times_1 \boldsymbol{U}^{(1)} \times_2 \boldsymbol{U}^{(2)} \times_3 \cdots \times_Q \boldsymbol{U}^{(Q)} \tag{70}$$

in which $\boldsymbol{\beta} \in \mathbb{R}^{[R_1]\times[R_2]\times\cdots\times[R_Q]}$ is the core tensor and the factors $\boldsymbol{U}^{(q)} \in \mathbb{R}^{[I_q]\times[R_q]}$ are thin matrices with $R_q \leq I_q$ (usually $R_q \ll I_q$) for all $q \in [Q]$. Note that the set of constraints in (69) becomes redundant; to see this we start from the following parametrization of the $q-$mode unfolding:

$$M_q(\boldsymbol{\alpha}) = \boldsymbol{U}^{(q)} M_q(\boldsymbol{\beta}) \odot_{j\neq q} \boldsymbol{U}^{(j)} \tag{71}$$

in which $\odot_{j\neq q}\boldsymbol{U}^{(j)}$ denotes the Kronecker product of the set of matrices $\left\{\boldsymbol{U}^{(j)} \ : \ j \in [Q] \setminus q\right\}$, see Appendix B and (89), in particular. Now one has:

$$\operatorname{rank}\left(M_q(\boldsymbol{\alpha})\right) \leq \min\left\{\operatorname{rank}\left(\boldsymbol{U}^{(q)}\right), \operatorname{rank}\left(M_q(\boldsymbol{\beta}) \otimes_{j\neq q} \boldsymbol{U}^{(j)}\right)\right\} \leq R_q \tag{72}$$

where we used the fact that $\operatorname{rank}(\boldsymbol{AB}) \leq \min\{\operatorname{rank}(\boldsymbol{A}), \operatorname{rank}(\boldsymbol{B})\}$.

## 5.2 Problem Reformulation

Consider an optimization problem involving the nuclear norm of a matrix $\boldsymbol{X} \in \mathbb{R}^{[I_1]\times[I_2]}$ where, without loss of generality, we can assume that $I_2 \leq I_1$. A common approach is to consider the following variational characterization:

$$\|\boldsymbol{X}\|_1 = \min_{\boldsymbol{X}=\boldsymbol{AB}^\top} \frac{1}{2}\left(\|\boldsymbol{A}\|_2^2 + \|\boldsymbol{B}\|_2^2\right) \tag{73}$$

in which $\boldsymbol{A} \in \mathbb{R}^{[I_1]\times[I_2]}$ and $\boldsymbol{B} \in \mathbb{R}^{[I_2]\times[I_2]}$. When, additionally, the rank of the unknown $\boldsymbol{X}$ is upper-bounded by $R$, a convenient approach is to consider thin factors $\boldsymbol{A} \in \mathbb{R}^{[I_1]\times[R]}$ and $\boldsymbol{B} \in \mathbb{R}^{[I_2]\times[R]}$ see, for instance, [36, Section 5.3]. Here we consider this approach for each of the nuclear norms in (69). In particular, keeping into account (71) we can write

$$M_q(\boldsymbol{\alpha}) = \boldsymbol{A}^{(q)}\boldsymbol{B}^{(q)\top} \text{ where } \boldsymbol{A}^{(q)} = \boldsymbol{U}^{(q)} M_q(\boldsymbol{\beta}) \text{ and } \boldsymbol{B}^{(q)} = \otimes_{j\neq q}\boldsymbol{U}^{(j)} \ .$$

This, in turn, results into the following unconstrained heuristic for the original non-convex constrained optimization problem (69):

$$\min_{\boldsymbol{\beta},\boldsymbol{U}} \left\{ \tfrac{1}{2\lambda} \sum_{n\in[N]} \left(y_n - \mathcal{S}(\boldsymbol{\beta}, \boldsymbol{U}; n)\right)^2 + 1/2 \sum_{q\in[Q]} \left( \left\|\boldsymbol{U}^{(q)} M_q(\boldsymbol{\beta})\right\|_2^2 + \prod_{j\neq q} \left\|\boldsymbol{U}^{(j)}\right\|_2^2 \right) \right\} \ , \tag{74}$$

in which $\mathcal{S}(\boldsymbol{\beta}, \boldsymbol{U}; n) := \boldsymbol{\beta} \times_1 \left(\boldsymbol{F}_{n\cdot}^{(1)}\boldsymbol{U}^{(1)}\right) \times_2 \cdots \times_Q \left(\boldsymbol{F}_{n\cdot}^{(Q)}\boldsymbol{U}^{(Q)}\right)$ is the empirical evaluation functional and we used the fact that $\left\| \odot_{j\neq q} \boldsymbol{U}^{(j)}\right\| = \prod_{j\neq q} \left\|\boldsymbol{U}^{(j)}\right\|$.

### 5.3 Block Descent Approach

Problem (74) involves finding the core tensor $\boldsymbol{\beta}$ as well as factors $\boldsymbol{U}^{(1)}, \boldsymbol{U}^{(2)}, \ldots, \boldsymbol{U}^{(Q)}$. It is not difficult to see that optimizing over each of these unknowns, conditioned on the rest of the unknowns being fixed, results into a convex quadratic problem. This suggests a simple block descent algorithm. Let $V_q$ denotes the $q-$mode vectorization operator given in Appendix A; define now $z^{(n)} \in \mathbb{R}^{R_1 R_2 \cdots R_Q}$, $c^{(n,q)} \in \mathbb{R}^{R_q}$ and $\boldsymbol{T}^{(n,q)} \in \mathbb{R}^{R_q I_q \times R_q I_q}$ by:

$$z^{(n)} := V_1 \left( \left( \boldsymbol{F}_{n\cdot}^{(1)} \boldsymbol{U}^{(1)} \right)^\top \otimes \left( \boldsymbol{F}_{n\cdot}^{(2)} \boldsymbol{U}^{(2)} \right)^\top \otimes \cdots \otimes \left( \boldsymbol{F}_{n\cdot}^{(M)} \boldsymbol{U}^{(M)} \right)^\top \right) \tag{75}$$

$$c^{(n,q)} := \boldsymbol{\beta} \times_1 \left( \boldsymbol{F}_{n\cdot}^{(1)} \boldsymbol{U}^{(1)} \right) \times_2 \cdots \times_{q-1} \left( \boldsymbol{F}_{\cdot n}^{(q-1)} \boldsymbol{U}^{(q-1)} \right) \times_{q+1} \tag{76}$$

$$\left( \boldsymbol{F}_{\cdot n}^{(q+1)} \boldsymbol{U}^{(q+1)} \right) \times_{q+2} \cdots \times_Q \left( \boldsymbol{F}_{\cdot n}^{(Q)} \boldsymbol{U}^{(Q)} \right) \tag{77}$$

$$\boldsymbol{T}^{(n,q)} := \left( c^{(n,q)} c^{(n,q)\top} \right) \odot \left( \boldsymbol{F}_{n\cdot}^{(q)\top} \boldsymbol{F}_{n\cdot}^{(q)} \right) \ . \tag{78}$$

Additionally set $g_q := \prod_{j \neq q} \|\boldsymbol{U}^{(j)}\|_2^2$ and let $\boldsymbol{G}_f^{(q)}$ and $\boldsymbol{G}_b^{(q)}$ be permutation matrices satisfying[13]:

$$\boldsymbol{G}_f^{(q)} V_1(\boldsymbol{\beta}) = V_q(\boldsymbol{\beta}) \quad \text{and} \quad \boldsymbol{G}_b^{(q)} V_q(\boldsymbol{\beta}) = V_1(\boldsymbol{\beta}) \ . \tag{79}$$

For $\boldsymbol{\beta}$, the first order optimality condition gives the following system of linear equations:

$$A(\boldsymbol{U}; \lambda) V_1(\boldsymbol{\beta}) = b(\boldsymbol{U}) \tag{80}$$

where, denoting by $\boldsymbol{I}^{(q)}$ the $J_q \times J_q$ identity matrix with $J_q = \prod_{j \neq q} R_j$, we have:

$$A(\boldsymbol{U}; \lambda) := \sum_{n \in [N]} z^{(n)} z^{(n)\top} + \lambda \sum_{q \in [Q]} \boldsymbol{G}_b^{(q)} \big( (\boldsymbol{U}^{(q)\top} \boldsymbol{U}^{(q)}) \odot \boldsymbol{I}^{(q)} \big) \boldsymbol{G}_f^{(q)} \tag{81}$$

$$b(\boldsymbol{U}) := \sum_{n \in [N]} y_n z^{(n)} \ . \tag{82}$$

Likewise, for $\boldsymbol{U}^{(q)}$ and each $q \in [Q]$ we get:

$$A\big(\boldsymbol{\beta}, \boldsymbol{U}^{\setminus q}; \lambda\big) V_1\big(\boldsymbol{U}^{(q)}\big) = b\big(\boldsymbol{\beta}, \boldsymbol{U}^{\setminus q}\big) \tag{83}$$

in which

$$A\big(\boldsymbol{\beta}, \boldsymbol{U}^{\setminus q}; \lambda\big) := \sum_{n \in [N]} \boldsymbol{T}^{(n,q)} + \lambda \left( M_q(\boldsymbol{\beta}) (M_q(\boldsymbol{\beta}))^\top + g_q \boldsymbol{I}_\ddagger^{(q)} \right) \odot \boldsymbol{I}_\dagger^{(q)} \tag{84}$$

$$b\big(\boldsymbol{\beta}, \boldsymbol{U}^{\setminus q}\big) := \sum_{n \in [N]} y_n c^{(n,q)} \odot \big( \boldsymbol{F}_{n\cdot}^{(q)} \big)^\top \tag{85}$$

and $\boldsymbol{I}_\dagger^{(q)}$ (respectively $\boldsymbol{I}_\ddagger^{(q)}$) is a $I_q \times I_q$ (respectively $R_q \times R_q$) identity matrix. Summing up we have the simple procedure stated in Algorithm 1, termed Mlrank-SNN. Each update in $\beta$ requires to solve a square linear system of size $R_1 R_2 \cdots R_Q$; each update in in $\boldsymbol{U}^{(q)}$, in turn, requires the solution of a square linear system of size $R_qI_q$. The computational complexity is therefore influenced by a number of factors. It depends upon the choice of upper-bound $(R_1, R_2, \ldots, R_Q)$. Additionally it is clear that being able to factorize the kernel matrix $\boldsymbol{K}^{(q)}$ in (55) into thin matrices ($I_q \ll N$), significantly speeds up calculations. Finally note that $N$, the number of training observations, enters linearly. It determines the number of rank-1 matrices (respectively, vectors) to sum up in (81) and (84) (respectively, (82) and (85)).

[13] Note that $\boldsymbol{G}_f^{(1)} = \boldsymbol{G}_b^{(1)} = \boldsymbol{I}$.

---

**Algorithm 1:** (Mlrank-SNN)

---

**Input:** $\lambda > 0$

**Output:** $\boldsymbol{\alpha} = \boldsymbol{\beta} \times_1 \boldsymbol{U}^{(1)} \times_2 \boldsymbol{U}^{(2)} \times_3 \cdots \times_Q \boldsymbol{U}^{(Q)}$

```
initialization: fix β_(0), U_(0)^(1), U_(0)^(2), ..., U_(0)^(Q)
k ← 0
repeat
    find the solution β_(k+1) to A(U_(k); λ)V_1(β) = b(U_(k)) in (80)
    U_(k+1) ← U_(k)
    for q ∈ [Q] do
        find the solution U_(k+1)^(q) to A(β_(k+1), U_(k+1)^{\q}; λ)V_1(U^(q)) = b(U_(k+1)^{\q}) in (83)
until stopping criterion met
β ← β_(k+1), U ← U_(k+1)
```

---

# 6 Experiments

## 6.1 Low Multilinear Rank Tensor Product Function

Our first test case arises from the low multilinear rank function of Section 3.1.2. We randomly draw input observations $x$ from the uniform distribution in $[0, 2\pi] \times [0, 2\pi] \times [0, 2\pi]$ and generate the corresponding outputs according to the model:

$$y = \boldsymbol{f}(x) + \epsilon \ .$$

In the latter, $\boldsymbol{f}$ is the function in (17) satisfying $\mathrm{mlrank}(\boldsymbol{f}) = (2, 3, 3)$ and $\epsilon$ is a zero-mean random variable with variance $\sigma^2$. We generated 3000 observations and used $N$ for training and the rest for testing. Within Mlrank-SNN we considered $(10, 10, 10)$ as upper-bound on the multilinear rank. In all the cases we terminated the algorithm when the iteration counter reached 100 or when the relative decrease in the objective function was smaller than $10^{-3}$. The approach was tested against a kernel-based learning scheme based on quadratic regularization, namely Least Squares Support Vector machine for Regression (LS-SVR) [48] implemented in LS-SVMlab [12]. Our goal is to assess multilinear spectral regularization against the classical approach based on the RKHS norm. Therefore we used exactly the same kernel function within the two procedures. In particular, we used the Gaussian-RBF kernel and performed model selection in LS-SVR via 10-fold cross-validation. The optimal kernel width in LS-SVR was then used also within Mlrank-SNN. The selection of regularization parameter within Mlrank-SNN was performed according to 5-fold cross-validation. The mean and standard deviation (in parenthesis) of the Mean Squared Error (MSE) obtained on the test set over 10 Monte Carlo runs is reported in Table 2 for different values of $N$ and two noise levels. In Figure 3 we reported the box-plot of the singular values of the different mode unfolding obtained from the different Monte Carlo runs. The plots refer to the noiseless case with $N = 300$. They show that Mlrank-SNN with cross-validated regularization parameter is able to capture the underlying multilinear rank.

Table 2: MSE for the Low Multilinear Rank function test case

$\sigma = 0$

| | $N$ | | |
|---|---|---|---|
| | 300 | 600 | 900 |
| LS-SVR | 1.765 (0.069) | 0.550 (0.071) | 0.134 (0.054) |
| Mlrank-SNN | 0.015 (0.006) | 0.008 (0.004) | 0.000 (0.000) |

$\sigma = 1$

| | $N$ | | |
|---|---|---|---|
| | 300 | 600 | 900 |
| LS-SVR | 3.395 (0.242) | 2.505 (0.116) | 2.050 (0.065) |
| Mlrank-SNN | 1.661 (0.079) | 1.493 (0.056) | 1.422 (0.165) |

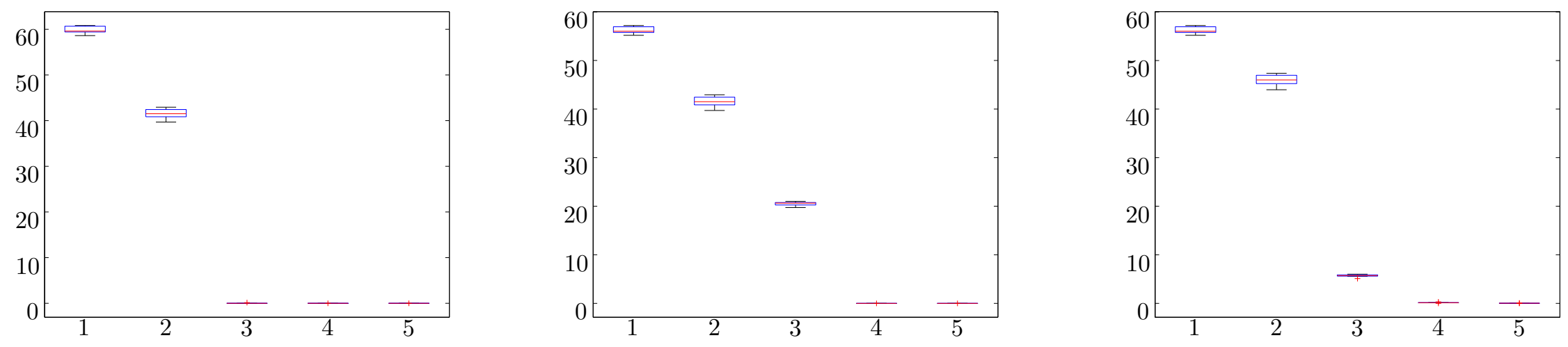

(a) 1−mode unfolding (b) 2−mode unfolding (c) 3−mode unfolding

Figure 3: Boxplots of the first five singular values for the estimated models.

## 6.2 Learning Preferences with Multimodal Side Information

Our second set of experiments relates to Section 3.3.3. With reference to Figure 1, we generated a $60 \times 60 \times 60$ tensor; the generic entry of the tensor, indexed by $\boldsymbol{i} = (i_1, i_2, i_3)$, is associated with a set of three vectors of attributes, one per mode: $g^{(1)}_{i_1}$, $g^{(2)}_{i_2}$ and $g^{(3)}_{i_3}$. Each of such vectors is drawn from a ten-dimensional normal distribution. The model for the preferences was taken to be a function of $\mathcal{H}_1 \otimes \mathcal{H}_2 \otimes \mathcal{H}_3$ where for any $m \in [3]$, $\mathcal{H}_m$ is the RKHS associated to the linear kernel $k^{(m)}(x, y) = \langle x, y \rangle$. Specifically we let:

$$\boldsymbol{g}(\boldsymbol{i}) = \sum_{n_1,n_2,n_3 \in [20]} \gamma_{n_1 n_2 n_3} \langle \tilde{g}^{(1)}_{n_1}, g^{(1)}_{i_1} \rangle \langle \tilde{g}^{(2)}_{n_2}, g^{(2)}_{i_2} \rangle \langle \tilde{g}^{(3)}_{n_3}, g^{(3)}_{i_3} \rangle + \epsilon(\boldsymbol{i}) \tag{86}$$

where $\epsilon(\boldsymbol{i})$ is a zero-mean normal Gaussian variable with variance $\sigma^2 = 0.01$, $\boldsymbol{\gamma}$ is a randomly generated tensor with multilinear rank $(2, 2, 2)$ and for any $m \in [3]$, $\left\{ \tilde{g}^{(m)}_{n_m} : n_m \in [20] \right\}$ is an independent set of vectors drawn from a ten-dimensional normal distribution. The indices for the $N$ preferences used for training were taken within the set $\{\boldsymbol{i} \in [60] \times [60] \times [60] \ : \quad i_m \leq 50, \ m \in [3]\}$; testing preferences were taken in the complementary set. Note that this ensures that the model is tested on previously unseen users and/or activities and/or locations. Following the discussion in Section 3.3.3, in this setting we can regard the task of learning preferences as the task of learning a (nonlinear) regression function in $R^{30}$. Therefore we again use LS-SVR (with Gaussian RBF-kernel) as a baseline strategy. Alternatively we use Mlrank-SNN with an upper-bound $(10, 10, 10)$ on the multilinear rank; kernel matrices per mode were constructed on training data based on the linear kernel. The MSE obtained on test preferences are reported in Table 3.

Table 3: MSE ($\times 10^{-3}$) for learning preferences with multimodal side information

| | $N$ | | | |
|---|---|---|---|---|
| | 625 | 1250 | 2500 | 5000 |
| LS-SVR | 1041.4 (6.5) | 866.4 (16.4) | 316.0 (12.4) | 63.2 (3.3) |
| Mlrank-SNN | 12.6 (0.25) | 11.2 (0.19) | 11.0 (0.20) | 10.8 (0.17) |

## 6.3 Multilinear Multitask Learning

Our third set of experiments relates to Section 3.3.2 where we have shown that a kernel-based view enables for (nonlinear) extensions of existing multilinear multitask learning models. To illustrate the idea we focus on the shoulder pain dataset [32]. This dataset contains video clips of the faces of people who suffer from shoulder pain. For each frame of the video, the facial expression is described by a set of 132 attributes (2D positions of 66 anatomical points). Each video is labelled frame by frame according to the physical activation of different set of facial muscles, encoded by the Facial Action Coding System [16]. This system defines a set of Action Units (AU) which refer to a contraction or relaxation of a determined set of muscles. As in [39] we aim at recognizing the AU intensity level of 5 different AUs for each of the 5 different patients. It was shown that applying the multilinear multitask learning approach proposed in [39] leads to significantly outperforming a number of alternative approaches. Here we test their algorithm based on the sum of nuclear norms (MLMTL-C) against Mlrank-SNN with an upper-bound $(10, 10, 10)$ on the multilinear rank. Within this approach we use either the kernel in (39) (lin-Mlrank-SNN), or the kernel in (40) (RBF-Mlrank-SNN). We used cross-validation to perform model selection within all the algorithms. As an additional baseline strategy we consider LS-SVR (with Gaussian RBF-kernel). In the latter case we learn a regression model where the 2 task indicators are adjoined to the vector of 132 features per frame. The rationale behind this choice is simple: since the Gaussian RBF-kernel is *universal* [47], one can learn an accurate model provided that a sufficient number of observations is given. The MSE obtained on test preferences are reported in Table 4.

Table 4: MSE for multilinear multitask learning

| | $N$ | | |
|---|---|---|---|
| | 200 | 300 | 400 |
| MLMTL-C | 0.366 (0.120) | 0.225 (0.045) | 0.189 (0.058) |
| LS-SVR | 0.434 (0.039) | 0.373 (0.030) | 0.291 (0.016) |
| LIN-MLRANK-SNN | 0.332 (0.094) | 0.230 (0.052) | 0.194 (0.045) |
| RBF-MLRANK-SNN | 0.272 (0.087) | 0.183 (0.026) | 0.156 (0.016) |

| | $N$ | | |
|---|---|---|---|
| | 500 | 600 | 700 |
| MLMTL-C | 0.165 (0.045) | 0.129 (0.031) | 0.109 (0.020) |
| LS-SVR | 0.254 (0.023) | 0.224 (0.017) | 0.200 (0.030) |
| LIN-MLRANK-SNN | 0.178 (0.067) | 0.145 (0.023) | 0.126 (0.013) |
| RBF-MLRANK-SNN | 0.147 (0.018) | 0.126 (0.010) | 0.118 (0.010) |

# 7 Conclusion

We have studied the problem of learning tensor product functions based on a novel class of regularizers, termed multilinear spectral penalties. We have shown that this framework comprises existing problem formulations as well as novel extensions. The approach relies on the generalization to the functional setting of a number of mathematical tools for finite dimensional tensors as well as a novel representer theorem. From a practical perspective the methodology involves finding a finite dimensional tensor; to cope with this task one could use a number of existing algorithms such as those developed for the sum of nuclear norms approach. As an additional contribution we have given a simple block descent algorithm. The latter tackles problems where the sum of nuclear norms of the functional unfoldings is combined with an upper bound on the multilinear rank. We have finally shown in experiments the usefulness of the proposed extensions.

## Acknowledgements

We thank Andreas Argyriou for fruitful discussions and Bernardino Romera-paredes for kindly providing the code for MLMTL-C. The scientific responsibility is assumed by its authors. Research supported by Research Council KUL: GOA/10/09 MaNet, PFV/10/002 (OPTEC); CIF1 STRT1/08/23; Flemish Government: IOF: IOF/KP/SCORES4CHEM, FWO: projects: G.0588.09 (Brain-machine), G.0377.09 (Mechatronics MPC), G.0377.12 (Structured systems), G.0427.10N (EEG-fMRI), IWT: projects: SBO LeCoPro, SBO Climaqs, SBO POM, EUROSTARS SMART iMinds 2013, Belgian Federal Science Policy Office: IUAP P7/19 (DYSCO, Dynamical systems, control and optimization, 2012-2017), EU: FP7-EMBOCON (ICT-248940), FP7-SADCO (MC ITN-264735), ERC ST HIGHWIND (259 166), ERC AdG A-DATADRIVE-B (290923). COST: Action ICO806: IntelliCIS.

# A Vector and Matrix Unfolding for Finite Dimensional Tensors

In this section we assume a finite dimensional tensor $\boldsymbol{\alpha} \in \mathbb{R}^{[I_1]\times[I_2]\times\cdots\times[I_M]}$ and we detail the practical implementation of the $m-$mode unfolding operator (Section 2.3); additionally we introduce the notion of $m-$mode vectorization operator used within Algorithm 1. In general, to implement unfolding operators one needs to fix one way to map a multi-index $\boldsymbol{i}$ into a single index $j$. In the following we assume that a bijection $\kappa : \boldsymbol{i} \mapsto j$ is given[14]. We define the 1-mode vector unfolding by:

$$[V_1(\boldsymbol{\alpha})]_{\kappa(i_1,i_2,i_3,\cdots,i_M)} := \boldsymbol{\alpha}_{i_1 i_2 i_3 \cdots i_M} \ .$$

Then for $1 < m \leq M$ the $m-$mode vector unfolding is defined by:

$$V_m(\boldsymbol{\alpha}) := V_1\big(\boldsymbol{\alpha}^{(m)}\big)$$

where

$$\begin{array}{rrcl} \boldsymbol{\alpha}^{(m)} : & [I_m]\times[I_2]\times\cdots\times[I_{m-1}]\times[I_1]\times[I_{m+1}]\times\cdots\times[I_M] & \to & \mathbb{R} \\ & (i_m, i_2, \cdots, i_{m-1}, i_1, i_{m+1}, \cdots, i_M) & \mapsto & \boldsymbol{\alpha}_{i_1 i_2 \cdots i_M} \end{array}$$

that is, $\boldsymbol{\alpha}^{(m)}$ is obtained by permuting[15] the 1st and $m$th dimension. Similarly, the 1-mode matrix unfolding of Section 2.3 can be equivalently defined, for finite dimensional tensors, by:

$$[M_1(\boldsymbol{\alpha})]_{i_1 \kappa(i_2,i_3,\cdots,i_M)} := \boldsymbol{\alpha}_{i_1 i_2 i_3 \cdots i_M} \ .$$

Note that if $\boldsymbol{\alpha} = f^{(1)} \otimes f^{(2)} \otimes \cdots \otimes f^{(M)}$, in particular, one has:

$$[M_1(\boldsymbol{\alpha})]_{i_1 \kappa(i_2,i_3,\cdots,i_M)} := f^{(1)}_{i_1} \otimes \left(f^{(2)} \otimes \cdots \otimes f^{(M)}\right)_{i_2 i_3 \cdots i_M} \ .$$

For $m > 1$ we now have:

$$M_m(\boldsymbol{\alpha}) := M_1\big(\boldsymbol{\alpha}^{(m)}\big) \ .$$

Note that, in the context of this paper, Proposition 2 makes it unnecessary to implement $\beta-$mode unfolding for a non-singleton $\beta$.

# B Partial Kronecker Product and $m-$mode Product

For $m \in \beta \subseteq [M]$ let $\mathcal{H}_m$ and $\mathcal{G}_m$ be HSs and consider the linear operator $A^{(m)} : \mathcal{H}_m \to \mathcal{G}_m$; the *Kronecker product* $\odot_{m\in\beta} A^{(m)}$, also denoted by $\boldsymbol{A}^{(\beta)}$, is defined for a rank-1 tensor by:

$$\begin{array}{rrcl} \boldsymbol{A}^{(\beta)} : & \otimes_{m\in\beta} \ \mathcal{H}_m & \to & \otimes_{m\in\beta}\mathcal{G}_m \\ & \otimes_{m\in\beta} \ f^{(m)} & \mapsto & \otimes_{m\in\beta}\big(A^{(m)} f^{(m)}\big) \ . \end{array} \quad (87)$$

[14]In MATLAB, for instance, the command `reshape(alpha,prod(size(alpha)),1)` uses

$$\kappa(i_1, i_2, \cdots, i_M) := 1 + \sum_{k=1}^{M} (i_k - 1) J_k \text{ where } J_k = \prod_{m=1}^{k-1} I_m \ .$$

[15]In MATLAB, for instance, $\boldsymbol{\alpha}^{(m)}$ can be obtained by `alpham=permute(alpha,[m setdiff(1:M,m)])`.

We call the Kronecker product partial whenever $\beta \subset [M]$; if $\beta = [M]$ we simply write $\boldsymbol{A}$. Consider a rank-1 function $\boldsymbol{f}$; it follows by the definitions of functional unfolding and Kronecker product that we have:

$$M_\beta(\boldsymbol{A}\boldsymbol{f}) = \big(\otimes_{m\in\beta} A^{(m)} f^{(m)}\big) \otimes \big(\otimes_{m\in\beta^c} A^{(m)} f^{(m)}\big) = \big(\boldsymbol{A}^{(\beta)} \odot \boldsymbol{A}^{(\beta^c)}\big) M_\beta(\boldsymbol{f}) \ . \tag{88}$$

We have yet another equivalent representation:

$$M_\beta(\boldsymbol{A}\boldsymbol{f}) = \boldsymbol{A}^{(\beta)} M_\beta(\boldsymbol{f}) \boldsymbol{A}^{(\beta^c)*} \ . \tag{89}$$

Indeed we have:

$$\begin{aligned}\big(\boldsymbol{A}^{(\beta)} M_\beta(\boldsymbol{f}) \boldsymbol{A}^{(\beta^c)*}\big)\boldsymbol{h} &\overset{\text{by (8)}}{=} \big(\boldsymbol{A}^{(\beta)} \big(\boldsymbol{f}^{(\beta)} \otimes \boldsymbol{f}^{(\beta^c)}\big) \boldsymbol{A}^{(\beta^c)*}\big)\boldsymbol{h} \overset{\text{by (6)}}{=} \big\langle \boldsymbol{f}^{(\beta^c)}, \boldsymbol{A}^{(\beta^c)*}\boldsymbol{h}\big\rangle_{\beta^c} \boldsymbol{A}^{(\beta)} \boldsymbol{f}^{(\beta)} \\ &= \big\langle \boldsymbol{A}^{(\beta^c)}\boldsymbol{f}^{(\beta^c)}, \boldsymbol{h}\big\rangle_{\beta^c} \boldsymbol{A}^{(\beta)} \boldsymbol{f}^{(\beta)} = \big(\big(\boldsymbol{A}^{(\beta)} \odot \boldsymbol{A}^{(\beta^c)}\big) M_\beta(\boldsymbol{f})\big)\boldsymbol{h} = \big(M_\beta(\boldsymbol{A}\boldsymbol{f})\big)\boldsymbol{h} \ . \end{aligned} \tag{90}$$

The (functional) $m-$mode product between $\boldsymbol{f}$ and $A^{(m)}$, denoted by $\boldsymbol{f} \times_m A^{(m)}$, is defined by:

$$\boldsymbol{f} \times_m A^{(m)} := \big(I^{(1)} \odot I^{(2)} \odot \cdots \odot I^{(m-1)} \odot A^{(m)} \odot I^{(m+1)} \odot \cdots \odot I^{(M)}\big)\boldsymbol{f} \tag{91}$$

where we denoted by $I^{(n)}$ the identity operator on $\mathcal{H}_n$. Note that (91) implies the following elementary properties:

$$\text{for } n \neq m: \qquad \big(\boldsymbol{f} \times_m A^{(m)}\big) \times_n A^{(n)} = \big(\boldsymbol{f} \times_n A^{(n)}\big) \times_m A^{(m)} = \boldsymbol{f} \times_m A^{(m)} \times_n A^{(n)} \tag{92}$$

$$\big(\boldsymbol{f} \times_m A^{(m)}\big) \times_m B^{(m)} = \boldsymbol{\alpha} \times_m \big(B^{(m)} A^{(m)}\big) \ . \tag{93}$$

Note that, as a special case of (88), we have:

$$\boldsymbol{g} = \boldsymbol{f} \times_m A^{(m)} \quad \Leftrightarrow \quad M_m(\boldsymbol{g}) = \big(A^{(m)} \odot I^{\{m\}^c}\big) M_m(\boldsymbol{f}) = A^{(m)} M_m(\boldsymbol{f}) \tag{94}$$

where we denoted by $I^{\{m\}^c}$ the identity operator on $\boldsymbol{\mathcal{H}}_{\{m\}^c}$. As before, the Kronecker and $n-$mode product and their properties hold for general elements of $\boldsymbol{\mathcal{H}}$, which are a linear combination of rank-1 tensors.

# C Proofs

## C.1 Proof of Proposition 1

Recall the definition of $e^{(i_m)}$ given in (31). Note that $\boldsymbol{k}$ in (39) can be equivalently restated as:

$$\boldsymbol{k}\left((z, i_2, \cdots, i_M), (z', i'_2, \cdots, i'_M)\right) = \left\langle z \otimes e^{(i_2)} \otimes e^{(i_3)} \otimes \ldots \otimes e^{(i_M)}, z' \otimes e^{(i'_2)} \otimes e^{(i'_3)} \otimes \ldots \otimes e^{(i'_M)} \right\rangle \tag{95}$$

where $z \otimes e^{(i_2)} \otimes e^{(i_3)} \otimes \ldots \otimes e^{(i_M)}$ is the representer of the evaluation functional in the tensor product space $\mathcal{H}_1 \otimes \mathbb{R}^{[I_2]} \otimes \mathbb{R}^{[I_3]} \otimes \cdots \otimes \mathbb{R}^{[I_M]}$, where $\mathcal{H}_1$ is a RKHS of functions on $\mathbb{R}^{I_1}$ with r.k. $k^{(1)}(z, z') = z^\top z'$. By definition of $\mathcal{D}_N$ in (38) it is enough to show that, for any $(x, y) \in \mathcal{D}_N$, we have $\langle \boldsymbol{f}, \boldsymbol{k}_x \rangle = w_t^\top z$ for $x = (z, i_2, \ldots, i_M)$ and $t = \kappa(\boldsymbol{i}) \in \left[\prod_{m=2}^M I_m\right]$, where $\boldsymbol{i}$ denotes the multi-index $(i_2, \ldots, i_M)$. This is shown as follows:

$$\begin{aligned}\langle \boldsymbol{f}, \boldsymbol{k}_z \rangle &\overset{\text{by (9)}}{=} \langle M_1(\boldsymbol{f}), M_1(\boldsymbol{k}_x) \rangle_{\text{HF}} \overset{\text{by (35) and (8)}}{=} \\ &\left\langle [w_1, w_2, \ldots, w_T], z \otimes \big(e^{(i_2)} \otimes e^{(i_3)} \otimes \cdots \otimes e^{(i_M)}\big) \right\rangle_{\text{HF}} = \\ &\sum_{\boldsymbol{j}} \big(e^{(i_2)} \otimes e^{(i_3)} \otimes \cdots \otimes e^{(i_M)}\big)_{\boldsymbol{j}} w_{\kappa(\boldsymbol{j})}^\top z = \big(e^{(i_2)} \otimes e^{(i_3)} \otimes \cdots \otimes e^{(i_M)}\big)_{\boldsymbol{i}} w_{\kappa(\boldsymbol{i})}^\top z = w_t^\top z \ . \end{aligned} \tag{96}$$

### C.2 Proof of Theorem 1

For any $q \in [Q]$, the functional unfolding $M_q$ is an isometry and therefore $\sum_{\beta\in\mathcal{B}} \|M_\beta(\boldsymbol{f})\|_2^2 = |\mathcal{B}|\|\boldsymbol{f}\|^2$. Additionally, for any $\beta \in \mathcal{B}$ one has $\|M_\beta(\boldsymbol{f})\|_2 = \|\boldsymbol{f}\| < \infty$; $M_\beta(\boldsymbol{f})$ is a Hilbert-Schmidt operator and therefore $\boldsymbol{H}_\mathcal{B} = \boldsymbol{\mathcal{H}}$. We can restate (43) as:

$$\min_{\boldsymbol{f}\in\boldsymbol{\mathcal{H}}} \left\{ \sum_{n\in[N]} l\left(y_n, \langle \boldsymbol{k}_{x_n}, \boldsymbol{f}\rangle\right) + \mu\|\boldsymbol{f}\|^2 \right\}$$

where $\mu := \lambda\,|\mathcal{B}|$. The theorem now follows from application of [41, Theorem 1].

### C.3 Proof of Proposition 2

Since $\mathcal{B}$ is a partition, we have that $\beta_i \cap \beta_j = \emptyset$ for $i \neq j$. We can identify a rank-1 $M-$th-order tensor $\boldsymbol{f} = f^{(1)} \otimes f^{(2)} \otimes \cdots \otimes f^{(M)} \in \boldsymbol{\mathcal{H}}$ with a rank-1 $Q$th-order tensor in $\boldsymbol{\mathcal{G}}$:

$$\iota(\boldsymbol{f}) = \left(\otimes_{m\in\beta_1} f^{(m)}\right) \otimes \left(\otimes_{m\in\beta_2} f^{(m)}\right) \otimes \cdots \otimes \left(\otimes_{m\in\beta_Q} f^{(m)}\right) = g^{(1)} \otimes g^{(2)} \otimes \cdots \otimes g^{(Q)} = \boldsymbol{g} \in \boldsymbol{\mathcal{G}} \ ; \quad (97)$$

in the latter, for any $q \in [Q]$, $g^{(q)} : x^{(\beta_q)} \mapsto \mathbb{R}$ is identified with $\otimes_{m\in\beta_q} f^{(m)}$. Moreover, we have:

$$M_{\beta_q}(\boldsymbol{f}) = \boldsymbol{f}^{(\beta_q)} \otimes \boldsymbol{f}^{(\beta_q^c)} = g^{(q)} \otimes \boldsymbol{g}^{(q^c)} = M_q(\boldsymbol{g}) \quad (98)$$

where $\boldsymbol{g}^{(q^c)} := \otimes_{j\in[Q]\setminus q} g^{(j)}$ is identified with $\otimes_{m\in\{\beta_q\}^c} f^{(m)}$. The correspondence of evaluations between $\boldsymbol{f} \in \boldsymbol{\mathcal{H}}$ and $\boldsymbol{g} := \iota(\boldsymbol{f})$ follows from the definition of rank-1 functions in (1). By the reproducing properties in $\boldsymbol{\mathcal{H}}$ and $\boldsymbol{\mathcal{G}}$ we have:

$$\langle \boldsymbol{f}, \boldsymbol{k}_x \rangle = \boldsymbol{f}\left(x^{(1)}, x^{(2)}, \ldots, x^{(M)}\right) = \boldsymbol{g}\left(x^{(\beta_1)}, x^{(\beta_2)}, \ldots, x^{(\beta_Q)}\right) = \langle \boldsymbol{g}, \iota(\boldsymbol{k}_x) \rangle \ . \quad (99)$$

Equations (98) and (99) imply now that $\mathcal{P}_\lambda^{\mathcal{H}}(\boldsymbol{f}) = \mathcal{P}_\lambda^{\mathcal{G}}(\iota(\boldsymbol{f}))$ for any $\boldsymbol{f} \in \boldsymbol{\mathcal{H}}$; likewise, for any $\boldsymbol{g} \in \boldsymbol{\mathcal{G}}$, one has $\mathcal{P}_\lambda^{\mathcal{G}}(\boldsymbol{g}) = \mathcal{P}_\lambda^{\mathcal{H}}(\iota^{-1}(\boldsymbol{g}))$. The equalities hold true, in particular, at any solution of the respective problems. Finally, equation (52) follows from $\langle \boldsymbol{g}, \iota(\boldsymbol{k}_x)\rangle = \langle \iota^{-1}(\boldsymbol{g}), \boldsymbol{k}_x \rangle$.

### C.4 Proof of Lemma 1

For an arbitrary $\beta \subseteq [Q]$ we define the Kronecker product $\boldsymbol{\Pi}^{(\beta)} : \otimes_{q\in\beta} g^{(q)} \mapsto \otimes_{q\in\beta}\left(\Pi_q g^{(q)}\right)$ and write $\boldsymbol{\Pi}$ to indicate $\boldsymbol{\Pi}^{([Q])}$. Note that, by construction, $\boldsymbol{\Pi}^{(\beta)}$ is a projection operator onto a closed linear subspace. Now we have:

$$\sigma_r\left(M_p\left(\boldsymbol{g} \times_1 \Pi_1 \times_2 \Pi_2 \times_3 \cdots \times_Q \Pi_Q\right)\right) = \sigma_r\left(M_p\left(\boldsymbol{\Pi}\,\boldsymbol{g}\right)\right) \overset{\text{by}(89)}{=} \sigma_r\left(\boldsymbol{\Pi}^p M_p\left(\boldsymbol{g}\right) \boldsymbol{\Pi}^{\{p\}^c}\right) \overset{(\star)}{\leq}$$

$$\sigma_r\left(M_p\left(\boldsymbol{g}\right) \boldsymbol{\Pi}^{\{p\}^c}\right) = \sigma_r\left(\boldsymbol{\Pi}^{\{p\}^c} M_p\left(\boldsymbol{g}\right)^*\right) \overset{(\star)}{\leq} \sigma_r\left(M_p\left(\boldsymbol{g}\right)^*\right) = \sigma_r\left(M_p\left(\boldsymbol{g}\right)\right) \quad (100)$$

where $(\star)$ follows from application of [1, Lemma 7].

### C.5 Proof of Theorem 2

By assumption there exists $\bar{\boldsymbol{f}} \in \arg\min_{\boldsymbol{f}\in \boldsymbol{H}_{\mathcal{B}}} \mathcal{P}^{\mathcal{H}}_{\lambda}(\boldsymbol{f})$; by Proposition 2, in turn, we have:

$$\bar{\boldsymbol{g}} := \iota(\bar{\boldsymbol{f}}) \in \arg\min_{\boldsymbol{g}\in \boldsymbol{G}_{\mathcal{B}}} \mathcal{P}^{\mathcal{G}}_{\lambda}(\boldsymbol{g})$$

where $\boldsymbol{G}_{\mathcal{B}} \subset \boldsymbol{\mathcal{G}} = \otimes_{q\in[Q]}\mathcal{G}_q$ and, for any $q \in [Q]$, $\mathcal{G}_q$ is the tensor product space $\otimes_{m\in\beta_q}\mathcal{H}_m$ with reproducing kernel $k^{(\beta_q)}$. Since for any $q \in [Q]$, the functional unfolding $M_q$ is an isometry and $\bar{\boldsymbol{g}} \in \boldsymbol{\mathcal{G}}$, one has $\|M_q(\bar{\boldsymbol{g}})\|_2 = \|\bar{\boldsymbol{g}}\| < \infty$; therefore $M_q(\bar{\boldsymbol{g}})$ is a Hilbert-Schmidt operator. For any $q \in [Q]$, denote by $\Pi_q$ the projection operator onto $\mathcal{V}_q$. For a set $\gamma = \{\gamma_1, \gamma_2, \ldots, \gamma_P\} \subseteq [Q]$ define the operator

$$P(\bar{\boldsymbol{g}}; \gamma) := \bar{\boldsymbol{g}} \times_{\gamma_1} \Pi_{\gamma_1} \times_{\gamma_2} \Pi_{\gamma_2} \times_{\gamma_3} \cdots \times_{\gamma_P} \Pi_{\gamma_P} \ .$$

Recall from Appendix B the definition of Kronecker product. For any $\boldsymbol{g} \in \boldsymbol{\mathcal{G}}$ we have:

$$\langle \iota(\boldsymbol{k}_{x_n}), \boldsymbol{g}\rangle = \left\langle \boldsymbol{k}^{(\beta_1)}_{x_n} \otimes \boldsymbol{k}^{(\beta_1^c)}_{x_n}, M_1(\boldsymbol{g})\right\rangle_{\mathrm{HF}} = \left\langle \Pi_1 \odot I^{\{1\}^c}\big(\boldsymbol{k}^{(\beta_1)}_{x_n} \otimes \boldsymbol{k}^{(\beta_1^c)}_{x_n}\big), M_1(\boldsymbol{g})\right\rangle_{\mathrm{HF}} = \left\langle \boldsymbol{k}^{(\beta_1)}_{x_n} \otimes \boldsymbol{k}^{(\beta_1^c)}_{x_n}, \Pi_1 \odot I^{\{1\}^c}\big(M_1(\boldsymbol{g})\big)\right\rangle_{\mathrm{HF}} \overset{\text{by}(94)}{=} \langle \iota(\boldsymbol{k}_{x_n}), \boldsymbol{g} \times_1 \Pi_1\rangle = \langle \iota(\boldsymbol{k}_{x_n}), P(\boldsymbol{g}; [1])\rangle \ . \quad (101)$$

With the same rationale we obtain:

$$\langle \iota(\boldsymbol{k}_{x_n}), P(\boldsymbol{g}; [1])\rangle = \left\langle \boldsymbol{k}^{(\beta_2)}_{x_n} \otimes \boldsymbol{k}^{(\beta_2^c)}_{x_n}, M_2(P(\boldsymbol{g}; [1]))\right\rangle_{\mathrm{HF}} = \cdots = \langle \iota(\boldsymbol{k}_{x_n}), P(\boldsymbol{g}; [2])\rangle \ .$$

Iterating the same procedure we see that, for any $n \in [N]$, we have:

$$\langle \iota(\boldsymbol{k}_{x_n}), \boldsymbol{g}\rangle = \cdots = \langle \iota(\boldsymbol{k}_{x_n}), P(\boldsymbol{g}; [Q])\rangle = \langle \iota(\boldsymbol{k}_{x_n}), \boldsymbol{g} \times_1 \Pi_1 \times_2 \Pi_2 \times_3 \ldots \times_Q \Pi_Q\rangle \quad (102)$$

which shows that $\bar{\boldsymbol{g}}$ and $P(\bar{\boldsymbol{g}}; [Q])$ have the same empirical evaluations. Now by Lemma 1 and the fact that $s$ follows the assumptions of Definition 1:

$$\Gamma\left(M_q(P(\bar{\boldsymbol{g}}; [Q])), q\right) = \sum_{r\in\mathbb{N}} s\left(\sigma_r\big(M_q\big(P(\bar{\boldsymbol{g}}; [Q])\big)\big), r, q\right) \leq \sum_{r\in\mathbb{N}} s\left(\sigma_r\big(M_q(\bar{\boldsymbol{g}})\big), r, q\right) = \Gamma\left(M_q(\bar{\boldsymbol{g}}), q\right) \quad \forall q \in [Q] \ . \quad (103)$$

From (102) and (103) we conclude that:

$$\mathcal{P}_{\lambda}(\bar{\boldsymbol{g}}) \geq \mathcal{P}_{\lambda}\left(P(\bar{\boldsymbol{g}}; [Q])\right) \ . \quad (104)$$

Since $\bar{\boldsymbol{g}}$ is a solution to (51), $P(\bar{\boldsymbol{g}}; [Q])$ must be feasible; therefore from (104) we conclude that $\hat{\boldsymbol{g}} := P(\bar{\boldsymbol{g}}; [Q])$ must also be a solution. By Proposition 2, $\hat{\boldsymbol{f}}(x) := \iota^{-1}(\hat{\boldsymbol{g}}(x))$ is a solution to (47) and we have:

$$\hat{\boldsymbol{g}}\big(x^{(\beta_1)}, x^{(\beta_2)}, \ldots, x^{(\beta_Q)}\big) = \hat{\boldsymbol{f}}(x)\big(x^{(1)}, x^{(2)}, \ldots, x^{(M)}\big) \quad \forall \ x = (x^{(1)}, x^{(2)}, \ldots, x^{(M)}) \in \mathcal{X} \ . \quad (105)$$

By construction $\hat{\boldsymbol{g}}$ belongs to $\in \boldsymbol{\mathcal{V}}_N = \otimes_{q\in[Q]}\mathcal{V}_q$ which is a subspace of $\otimes_{q=1}^{Q}\mathcal{G}_q$ with dimension $I_1 I_2 \cdots I_Q$. Since for $q \in [Q]$ (45) is an orthonormal basis for $\mathcal{V}_q$, the set:

$$\left\{u^{(1)}_{i_1} \otimes u^{(2)}_{i_2} \otimes \cdots \otimes u^{(Q)}_{i_Q} \ : \ (i_1, i_2, \ldots, i_Q) \in [I_1] \times [I_2] \times \cdots \times [I_Q]\right\}$$

forms an orthonormal basis for $\boldsymbol{\mathcal{V}}_N$. Therefore there must be a (unique) tensor $\hat{\boldsymbol{\alpha}} \in \mathbb{R}^{[I_1]\times[I_2]\times\cdots\times[I_Q]}$ so that:

$$\hat{\boldsymbol{g}} = \sum_{i_1\in[I_1]}\sum_{i_2\in[I_2]}\cdots\sum_{i_Q\in[I_Q]} \hat{\boldsymbol{\alpha}}_{(i_1,i_2,\cdots,i_Q)} u^{(1)}_{i_1}\otimes u^{(2)}_{i_2}\otimes\cdots\otimes u^{(Q)}_{i_Q}\ , \tag{106}$$

which concludes the proof.

## C.6 Proof of Proposition 3

By Theorem C.5 we know that problem (47) reduces to find a solution to the finite dimensional problem

$$\min_{\boldsymbol{g}\in\boldsymbol{\mathcal{V}}_N} \sum_{n\in[N]} l\left(y_n, \langle\iota(\boldsymbol{k}_{x_n}),\boldsymbol{g}\rangle\right) + \lambda\sum_{q\in[Q]}\Gamma\left(M_q(\boldsymbol{g}),q\right)\ ; \tag{107}$$

where $\iota : \boldsymbol{\mathcal{H}} \to \boldsymbol{\mathcal{G}}$ denotes a vector space isomorphism. In turn, a solution to the latter admits the representation (106) and therefore it is only required to find the finite dimensional tensor $\hat{\boldsymbol{\alpha}} \in \mathbb{R}^{[I_1]\times[I_2]\times\cdots\times[I_Q]}$. From (106) notice that we have, for any $q\in[Q]$:

$$M_q(\hat{\boldsymbol{g}}) = \sum_{i_q\in[I_q]}\sum_{\boldsymbol{j}\in\boldsymbol{J}}\left(M_q(\hat{\boldsymbol{\alpha}})\right)_{i_q\boldsymbol{j}} u^{(q)}_{i_q}\otimes \boldsymbol{u}^{(q^c)}_{\boldsymbol{j}} \tag{108}$$

where $\boldsymbol{j}$ denotes a multi-index in $\boldsymbol{J} = \times_{m\neq q}[I_m]$; since $\left\{u^{(q)}_{i_q}\right\}$ and $\left\{\boldsymbol{u}^{(q^c)}_{\boldsymbol{j}}\right\}$ are orthonormal bases in the respective spaces, the non-zero singular values of $M_q(\hat{\boldsymbol{g}})$ are now the non-zero singular values of the matrix $M_q(\hat{\boldsymbol{\alpha}})$. Therefore, with reference to (107), we have:

$$\Gamma\left(M_q(\hat{\boldsymbol{g}}),q\right) = \Gamma\left(M_q(\hat{\boldsymbol{\alpha}}),q\right)\ \text{ for any } q\in[Q]\ . \tag{109}$$

In order to obtain an optimization problem in the parameter $\hat{\boldsymbol{\alpha}}$ it remains to restate the evaluation on the $n-$th input training data $\hat{\boldsymbol{g}}(x_n) = \langle\iota(\boldsymbol{k}_{x_n}),\hat{\boldsymbol{g}}\rangle$ as a function of $\hat{\boldsymbol{\alpha}}$. We begin by restating (106) in terms of evaluations of kernels centered at training data rather than orthonormal basis elements:

$$\hat{\boldsymbol{g}}(\cdot) = \sum_{n_1\in[N]}\sum_{n_2\in[N]}\cdots\sum_{n_Q\in[N]}\gamma_{n_1n_2\cdots n_Q}\boldsymbol{k}^{(\beta_1)}\left(x^{(\beta_1)}_{n_1},\cdot^{(\beta_1)}\right)\otimes$$
$$\boldsymbol{k}^{(\beta_2)}\left(x^{(\beta_2)}_{n_2},\cdot^{(\beta_2)}\right)\otimes\cdots\otimes\boldsymbol{k}^{(\beta_Q)}\left(x^{(\beta_Q)}_{n_Q},\cdot^{(\beta_Q)}\right)\ . \tag{110}$$

For any $q\in[Q]$, consider the Gram matrix $\boldsymbol{K}^{(q)}_{ij} = k^{(\beta_q)}(x_i,x_j)$ . From (45) we see that $\boldsymbol{K}^{(q)}$ has rank $I_q$; it therefore admits a factorization $\boldsymbol{K}^{(q)} = \boldsymbol{F}^{(q)}\boldsymbol{F}^{(q)\top}$ where $\boldsymbol{F}^{(q)}\in\mathbb{R}^{[N]\times[I_q]}$ is an arbitrary square root. Note that $\boldsymbol{F}^{(q)}_{ni_q}$ represent the coefficient of the function $k^{(\beta_q)}(x^{(\beta_q)}_n,\cdot)$ with respect to the $i_q$-th element of an orthonormal basis for the space (46). In light of this, we have:

$$\boldsymbol{k}^{(\beta_1)}\left(x^{(\beta_1)}_{n_1},\cdot\right)\otimes\boldsymbol{k}^{(\beta_2)}\left(x^{(\beta_2)}_{n_2},\cdot\right)\otimes\cdots\otimes\boldsymbol{k}^{(\beta_Q)}\left(x^{(\beta_Q)}_{n_Q},\cdot\right) =$$
$$\sum_{i_1\in[I_1]]}\sum_{i_2\in[I_2]}\cdots\sum_{i_Q\in[I_Q]}\boldsymbol{F}^{(1)}_{n_1i_1}\boldsymbol{F}^{(1)}_{n_2i_2}\cdots\boldsymbol{F}^{(Q)}_{n_Qi_Q}u^{(1)}_{i_1}(\cdot)\otimes u^{(2)}_{i_2}(\cdot)\otimes\cdots\otimes u^{(Q)}_{i_Q}(\cdot)\ . \tag{111}$$

Now replacing the latter into (110) we obtain:

$$\hat{\boldsymbol{g}} = \sum_{i_1\in[I_1]}\sum_{i_2\in[I_2]}\cdots\sum_{i_Q\in[I_Q]}\left(\sum_{n_1\in[N]}\sum_{n_2\in[N]}\cdots\sum_{n_Q\in[N]}\boldsymbol{\gamma}_{n_1n_2\cdots n_Q}\boldsymbol{F}^{(1)}_{n_1i_1}\boldsymbol{F}^{(1)}_{n_2i_2}\cdots\right.$$
$$\left.\boldsymbol{F}^{(Q)}_{n_Qi_Q}\right)u^{(1)}_{i_1}\otimes u^{(2)}_{i_2}\otimes\cdots\otimes u^{(Q)}_{i_Q} \quad (112)$$

which, compared to (106) yields:

$$\hat{\boldsymbol{\alpha}}_{i_1i_2\cdots i_Q} = \sum_{n_1\in[N]}\sum_{n_2\in[N]}\cdots\sum_{n_Q\in[N]}\boldsymbol{\gamma}_{n_1n_2\cdots n_Q}\boldsymbol{F}^{(1)}_{n_1i_1}\boldsymbol{F}^{(1)}_{n_2i_2}\cdots\boldsymbol{F}^{(Q)}_{n_Qi_Q}\ . \quad (113)$$

With reference to (107) the evaluation on the $n-$th input training data $\hat{\boldsymbol{g}}(x_n) = \langle\iota(\boldsymbol{k}_{x_n}), \hat{\boldsymbol{g}}\rangle$ is now obtained starting from (110) as:

$$\hat{\boldsymbol{g}}(x_n) = \sum_{n_1\in[N]}\sum_{n_2\in[N]}\cdots\sum_{n_Q\in[N]}\boldsymbol{\gamma}_{n_1n_2\cdots n_Q}\boldsymbol{K}^{(1)}_{n_1n}\boldsymbol{K}^{(2)}_{n_2n}\cdots\boldsymbol{K}^{(Q)}_{n_Qn} =$$
$$\sum_{n_1\in[N]}\sum_{n_2\in[N]}\cdots\sum_{n_Q\in[N]}\boldsymbol{\gamma}_{(n_1n_2\cdots n_Q)}\left(\sum_{i_1\in[I_1]}\boldsymbol{F}^{(1)}_{n_1i_1}(\boldsymbol{F}^{(1)\top})_{i_1n}\right)\cdots\left(\sum_{i_Q\in[I_Q]}\boldsymbol{F}^{(Q)}_{n_Qi_Q}(\boldsymbol{F}^{(Q)\top})_{i_Qn}\right) =$$
$$\sum_{i_1\in[I_1]}\sum_{i_2\in[I_2]}\cdots\sum_{i_Q\in[I_Q]}\left(\sum_{n_1\in[N]}\sum_{n_2\in[N]}\cdots\sum_{n_Q\in[N]}\boldsymbol{\gamma}_{n_1n_2\cdots n_Q}\boldsymbol{F}^{(1)}_{n_1i_1}\cdots\boldsymbol{F}^{(Q)}_{n_Qi_Q}\right)(\boldsymbol{F}^{(1)\top})_{i_1n}\cdots$$
$$(\boldsymbol{F}^{(Q)\top})_{i_Qn} \overset{\text{by (113)}}{=} \hat{\boldsymbol{\alpha}}\times_1\boldsymbol{F}^{(1)}_{n\cdot}\times_2\cdots\times_Q\boldsymbol{F}^{(Q)}_{n\cdot}\ . \quad (114)$$

From this and (109) we obtain problem (54).